\documentclass[10pt,journal,compsoc]{IEEEtran}

\ifCLASSOPTIONcompsoc
  \usepackage[nocompress]{cite}
\else
  \usepackage{cite}
\fi
\ifCLASSINFOpdf
\else
\fi

\usepackage[colorlinks,urlcolor=blue,linkcolor=blue,citecolor=blue]{hyperref}
\usepackage{soul}
\usepackage{multirow}
\usepackage{graphicx}
\usepackage{subfigure}
\usepackage{graphics}
\usepackage{amsmath}
\usepackage{amsfonts}
\usepackage{algorithm}
\usepackage{algorithmic}
\usepackage{multirow}
\usepackage{booktabs}
\usepackage{pifont}
\usepackage{color}
\usepackage{verbatim}
\usepackage{threeparttable}
\usepackage[dvipsnames]{xcolor}
\usepackage{pdfpages}
\usepackage{color}
\usepackage{framed}
\usepackage{enumitem}

\newcommand{\fm}[1]{\textcolor{black}{#1}}
\newcommand{\ff}[1]{\textcolor{black}{#1}}
\AtBeginDocument{%
  \providecommand\BibTeX{{%
    \normalfont B\kern-0.5em{\scshape i\kern-0.25em b}\kern-0.8em\TeX}}}

\begin{document}

\title{Fight Perturbations with Perturbations: Defending Adversarial Attacks\\ via Neuron Influence}

\author{Ruoxi Chen, Haibo Jin, Haibin Zheng, Jinyin Chen, Zhenguang Liu
\IEEEcompsocitemizethanks{\IEEEcompsocthanksitem This research was supported by the Zhejiang Provincial Natural Science Foundation (No. LDQ23F020001), National Natural Science Foundation of China (No. 62072406), Zhejiang Province Key R\&D Science and Technology Plan Project (No. 2022C01018).
\IEEEcompsocthanksitem R.~Chen and H.~Jin are with the College of Information Engineering at Zhejiang University of Technology, Hangzhou, 310023, China. (e-mail: 2112003149@zjut.edu.cn, 2112003035@zjut.edu.cn)
\IEEEcompsocthanksitem H.~Zheng and J.~Chen are with the Institute of Cyberspace Security and the College of Information Engineering at Zhejiang University of Technology, Hangzhou, 310023, China. (e-mail: haibinzheng320@gmail.com, chenjinyin@zjut.edu.cn) 
\IEEEcompsocthanksitem Z.~Liu is with the School of Cyber Science and Technology at Zhejiang University, Hangzhou, 310023, China. (e-mail: liuzhenguang2008@gmail.com) 
\IEEEcompsocthanksitem  Corresponding author: Jinyin Chen. e-mail: chenjinyin@zjut.edu.cn
}
\thanks{Manuscript received xx xx, 2023; revised xx xx, xxxx.}}

\markboth{IEEE TRANSACTIONS ON DEPENDABLE AND SECURE COMPUTING,~Vol.~xx, No.~xx, xxxx~2023}%
{Chen \MakeLowercase{\textit{et al.}}: Fight Perturbations with Perturbations: Defending Adversarial Attacks via Neuron Influence}

\IEEEtitleabstractindextext{%
\begin{abstract}
The vulnerabilities of deep learning models towards adversarial attacks have attracted increasing attention, especially when models are deployed in security-critical domains. 
Numerous defense methods, 
including reactive and proactive ones, 
have been proposed for model robustness improvement. 
Reactive defenses, such as conducting transformations to remove perturbations,
usually fail to handle large perturbations.
The proactive defenses that involve retraining, 
suffer from the attack dependency and high computation cost.
In this paper, we consider defense methods from the general effect of adversarial attacks that take on neurons inside the model. 
We introduce the concept of neuron influence, 
which can quantitatively measure neurons' contribution to correct classification. 
Then, we observe that almost all attacks fool the model by suppressing neurons with larger influence and enhancing those with smaller influence. 
Based on this, we propose \emph{Neuron-level Inverse Perturbation} (NIP), a novel defense against general adversarial attacks. 
It calculates neuron influence from benign examples and then modifies input examples by generating inverse perturbations that can in turn strengthen neurons with larger influence and weaken those with smaller influence. Extensive experiments on benchmark datasets and models show that NIP outperforms the state-of-the-art methods in terms of (i) \emph{effective} - it shows better defense success rate ($\sim\!\!\times 1.45$) against 13 adversarial attacks; (ii) \emph{elastic} - it maintains better defense ($\sim\!\!\times 3.4$ in the worst case) on large perturbations; (iii) \emph{efficient} - it runs with only $\sim\!\!1/6$ time cost; (iv) \emph{extensible} - it can be applied to speaker recognition models and Baidu online image platforms. We further evaluate NIP against potential adaptive attacks and provide interpretable analysis for its effectiveness.
The code of NIP is open-sourced at \url{https://github.com/Allen-piexl/NIP-Neuron-level-Inverse-Perturbation}.
\end{abstract}

}

\maketitle
\IEEEdisplaynontitleabstractindextext
\IEEEpeerreviewmaketitle
\section{Introduction}
\IEEEPARstart{D}{eep} neural networks (DNNs) have demonstrated impressive capabilities in both academic and industrial areas, including speech recognition~\cite{ji2022asrtest, liu2021dialtest}, computer vision~\cite{yu2022automated,LiangLSLSW21}, etc. 
However, their vulnerabilities towards adversarial perturbations have raised public concern~\cite{zhao2021attack}. 
\fm{These well-crafted perturbations~\cite{GoodfellowSS14}, 
though often imperceptible to humans, 
pose grave a threat to mission-critical domains~\cite{rozsa2019facial,abdessalem2020automated,guo2022lirtest}. For instance, in the domain of autonomous driving, there is a notable case where Tesla vehicles were tricked into significantly increasing their speed~\cite{tesla}. This was achieved by a malicious attacker subtly altering a roadside speed limit sign, leading the car to accelerate by an additional 50 miles per hour. Therefore, it is crucial to enhance the robustness of DNNs against adversarial attacks to ensure safety and reliability in critical applications.} 

So far, intensive research has been devoted to improving DNNs' robustness, in two main directions: 
reactive and proactive defenses~\cite{mustafa2020deeply}. 
The former ones apply transformations to the input during the inference stage, 
while the latter ones modify the model structure or training procedure to mitigate attacks.
Early-proposed reactive strategies~\cite{dziugaite2016study,Xu0Q18,guo2018countering,liu2019feature}
attempt to remove perturbations by simple image transformations. 
But they cannot handle larger perturbations 
(e.g., $l_2$-norm of perturbation is larger than 0.6 on CIFAR-10~\cite{krizhevsky2009learning}). 
Besides, massive adversarial examples are required for choosing proper parameters.
Meanwhile, other reactive defenses are put forward from the perspective of feature distribution~\cite{sun2019adversarial, mustafa2019image}, which push the distribution of adversarial examples to approximate benign ones via mapping and projection in latent feature space. 
Their performances are limited by the precision of the mapping function and perturbation sizes. 

Furthermore, if full knowledge of model structures is given in advance, proactive defense can provide better model robustness. Several methods that develop model operations or optimization objectives~\cite{DhillonALBKKA18, xie2019feature}, have verified defense effectiveness. Adversarial training~\cite{GoodfellowSS14, mustafa2020deeply} that adds adversarial examples of specific attacks to the training dataset, is commonly adopted for defense. Since DNNs' decisions are determined by the nonlinear combination of each neuron, some adversarial training strategies are designed from the aspect of neuron activation values and weights~\cite{DhillonALBKKA18,zhang2019neuron, suri2020one, bai2020improving, zhang2020interpreting}. Connecting neuron activation with model robustness, recent promising methods~\cite{bai2020improving, zhang2020interpreting} assume the misclassification of adversarial examples can be attributed to abnormally activated neurons. But they rely on massive projected gradient descent (PGD)~\cite{Madry2018Towards} adversarial examples, which also limits effectiveness on general attacks. During training, they are also challenged by high computation costs and the drop of benign accuracy.


Based on the above analysis, 
existing defenses show limitations from three aspects: 
(i) unstable performance among different attacks; 
(ii) cannot adaptively adjust the defense strategy towards different perturbation sizes; 
(iii) rely on adversarial examples from specific attacks and require high time cost during retraining.

\begin{figure}[t]
\centering
    \subfigure[FGSM~\cite{GoodfellowSS14}]{
        \includegraphics[width=0.48\linewidth]{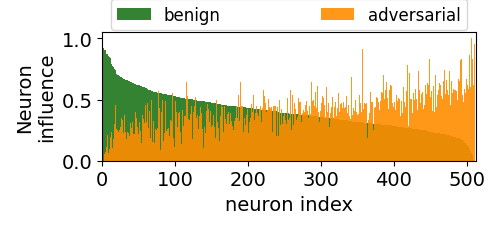}}
    \subfigure[DeepFool~\cite{moosavi2016deepfool}]{
        \includegraphics[width=0.48\linewidth]{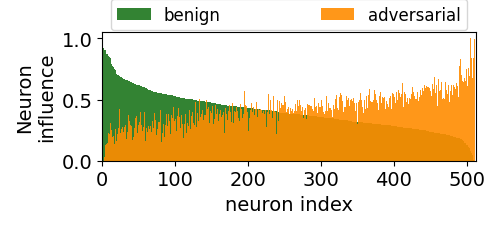}}\\
        \vspace{-3pt}
     \subfigure[Boundary~\cite{Brendel2018Boundary}]{
        \includegraphics[width=0.48\linewidth]{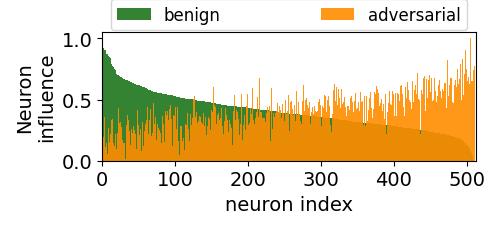}}
    \subfigure[PWA~\cite{SchottRBB19}]{
        \includegraphics[width=0.48\linewidth]{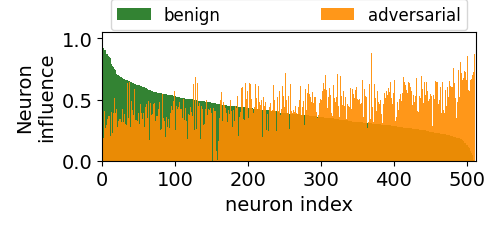}}   
\vspace{-13pt}
\caption{Changes of neuron influence on benign and adversarial examples on average. Dataset: CIFAR-10; model: VGG19~\cite{simonyan2014very}; layer: flatten layer with 512 neurons.}
\label{intro_attack}
\vspace{-16pt}
\end{figure}

\fm{To overcome the above challenges, 
we reconsider the defense mechanism from the consistency of different attacks and pay special attention to changes in neuron behaviors before and after attacks. }
Assuming that neurons in the model play different roles in classification, 
we introduce \emph{neuron influence}, 
which quantitatively measures the extent of neuron's contribution to correct predictions. 
It is calculated under the guidance of correct labels when the model is fed with benign examples. 
According to neuron influence, neurons in a certain layer are sorted by their influence values in descending order and divided into front neurons (the most class-relevant), 
tail neurons (the least class-relevant) and remaining neurons. 
We visualize the neuron influence of 500 benign and corresponding adversarial examples in Fig. \ref{intro_attack}, where influence values are normalized to [0,1]. \fm{Observations indicate that changes in neuron influence under different adversarial attacks display a consistent pattern: front neurons typically experience a reduction in influence, while the influence of tail neurons increases. This regularity suggests that general adversarial attacks commonly target these specific neurons to manipulate the behavior of neural networks, ultimately leading to misclassifications of adversarial examples.}

\fm{Interpretation of the results clearly suggests that effectively reversing the activation of neurons that have been maliciously targeted — either activated or inactivated — can significantly reduce or eliminate the influence of adversarial examples on the model. Based on the general pattern of different attacks, in this paper, we propose an attack-agnostic defense method \emph{Neuron-level Inverse Perturbation} (NIP). }
First, the neuron influence in the chosen layer is calculated by a few batches of benign examples, and then front and tail neurons that may be utilized by adversaries are identified. \fm{Utilizing these neurons, we generate inverse perturbations to modify the input value, such as image pixels and audio spectrum values, during the inference stage. Notably, this modification process does not require the expensive retraining of the model. To effectively handle adversarial examples that vary in perturbation size, we adaptively adjust the amplitude of the inverse perturbations based on the properties of each unknown input.}


The main contributions are summarized as follows.
\begin{itemize}[nolistsep, leftmargin=*]
    \item 
    We propose the concept of neuron influence, which quantitatively interprets
    DNNs’ misclassifications, i.e., adversarial attacks fool the model by enhancing the least class-relevant neurons with small influence and suppressing those with large influence.
    \item 
    We design an attack-agnostic defense method NIP, which generates inverse perturbations based on neuron influence to modify input values. It can mitigate or even eliminate the effect of general adversarial perturbations, regardless of perturbation sizes. 
    \item 
    Extensive experiments on various datasets and models have shown that NIP outperforms the state-of-the-art (SOTA) defense strategies against 13 attacks. NIP can also be applied to speaker recognition models and Baidu online platforms.
    \item 
    We further discuss adaptive attacks and provide analytical justification for effective defense. 
\end{itemize}

\section{Related Work\label{RWs}}
In this section, we review the related work and briefly summarize attack and defense methods used in the experiment. Besides, existing neuron-level approaches for DNNs in security domains are introduced as well. 
\subsection{Adversarial Attacks}
According to whether the attacker requires prior knowledge of the targeted model, adversarial attacks can be divided into white-box attacks and black-box attacks.

\subsubsection{White-box Attacks} 
White-box attacks mainly use gradient information for adversarial perturbation generation. Goodfellow et al.~\cite{GoodfellowSS14} proposed FGSM, adding perturbation along the direction where the gradient of the model changes the most. Based on it, basic iterative method (BIM)~\cite{Kurakin2017Adversarial} and momentum iterative FGSM (MI-FGSM) ~\cite{dong2018boosting} both generate smaller perturbations with better effectiveness. PGD~\cite{Madry2018Towards} does several iterations and projects the perturbation to the specified range, achieving stronger attack than previous ones. Papernot et al.~\cite{papernot2016limitations} proposed JSMA to calculate perturbations based on the critical pixels in the saliency map. Moosavi et al.~\cite{moosavi2016deepfool} gradually pushed the image within the classification boundary to the outside until the wrong predicted label occurred. Perturbations calculated by Moosavi et al.~\cite{moosavi2017universal} can fool the network on ``any" image with high probability, namely universal adversarial perturbations (UAP). Besides, Athalye et al. \cite{athalye2018synthesizing} presented Expectation Over Transformation (EOT) attack, for synthesizing examples that are adversarial over a chosen distribution of transformations in the real world.


\begin{table*}[t]
\centering
\caption{The characteristics of different neuron-level defenses.  }
\vspace{-8pt}
 \begin{threeparttable}
\resizebox{0.95\linewidth}{!}{
\begin{tabular}{llllllll}
\toprule
\textbf{Defenses} &
  \textbf{\begin{tabular}[c]{@{}l@{}}Defense\\ type\end{tabular}} &
  \textbf{\begin{tabular}[c]{@{}l@{}}Adv-\\ free?\end{tabular}} &
  \textbf{\begin{tabular}[c]{@{}l@{}}Re-\\ train?\end{tabular}} &
  \textbf{\begin{tabular}[c]{@{}l@{}}Class-\\ indep?\end{tabular}} &
  \textbf{Model knowledge} &
  \textbf{What's modified?} &
  \textbf{How to use neurons?} \\ \hline
SAP \cite{DhillonALBKKA18}              & Adversarial      & Yes & No  & Yes & All activation layer & Activation value & Randomly magnify or zero the activation value  \\
NS \cite{zhang2019neuron} & Adversarial      & Yes & No  & No  & All neurons          & Activation value & Zero the activation value                      \\
Suri et al. \cite{suri2020one}      & Adversarial      & No  & Yes & No  & All neurons          & Weights          & Reweight neuron weights and logits             \\
CAS \cite{bai2020improving}             & Adversarial      & No  & Yes & Yes & All neurons          & Weights          & Reweight channels based on their contributions \\
SNS \cite{zhang2020interpreting}  & Adversarial      & No  & Yes & Yes & All neurons          & Weights          & Suppress neuron activations                    \\
 Fine-pruning \cite{liu2018fine}     & Backdoor & /   & Yes &  /   &  All neurons          &  Activation value &  Prune neurons with low activation values       \\
 ABS \cite{liu2019abs}             &  Backdoor & /   &  No  & /   &  All neurons          &  /               &  Target abnormal neurons to detect backdoors    \\
 ANP \cite{wu2021adversarial}             &  Backdoor & /   &  Yes  & /   &  All neurons          &  Weights                &  Prune neurons sensitive to weight perturbations    \\
\textbf{NIP}     & Adversarial      & Yes & No  & Yes & One layer            & Examples         & Generate inverse perturbations                 \\
\bottomrule

  \vspace{-10pt}   
  \end{tabular}}
\begin{tablenotes}
        \footnotesize
       \item[] ``Adv-free'' means free of adversarial examples and ``Class-indep'' means class independent.
      \end{tablenotes}
      \end{threeparttable}
        \label{comparison}
      \vspace{-10pt}     
  \end{table*}

\subsubsection{Black-box Attacks} 
Black-box approaches fool the model by the output label or confidence, without specific parameters of models. Brendel et al.~\cite{Brendel2018Boundary} proposed boundary attack based on the model decision, which seeks smaller perturbations while staying adversarial. Su et al.~\cite{su2019one} modified only one pixel of the image for generating perturbations based on differential evolution. Andriushchenko et al.~\cite{andriushchenko2020square} designed a query-efficient attack algorithm based on model scores, achieving a higher success rate in untargeted setting. PWA~\cite{SchottRBB19} performs a binary search for finding effective adversarial while AUNA ~\cite{FoolboxAUNA} adds uniform noise and gradually increases the deviation until the model is fooled.

\fm{Besides, Croce et al. \cite{croce2020reliable} combined extensions of PGD attack with other complementary attacks. They proposed AutoAttack, which is parameter-free, computationally affordable and user-independent}.

\subsection{Defenses against Attacks}
Defense techniques could be mainly categorized into two types: reactive defense and proactive defense. 

\subsubsection{Reactive Defenses} 
Reactive methods perform pre-processing operations and transformations on the input images before inputting them to the target model~\cite{mustafa2020deeply}. Dziugaite et al.~\cite{dziugaite2016study} found that JPG compression could reverse the drop in classification accuracy of adversarial examples to a large extent. Xu et al.~\cite{Xu0Q18} reduced the search space available to an adversary by reduction of color bit depth and spatial smoothing. Guo et al.~\cite{guo2018countering} conducted input transformations such as total variance minimization to adversarial examples before feeding them into the model. Based on randomness, these methods are effective but hard to handle larger perturbations. Besides, Mustafa et al.~\cite{mustafa2019image} put forward super resolution as a defense mechanism, which projects adversarial examples back to the natural image manifold learned by models. Similarly, Sun et al.~\cite{sun2019adversarial} used sparse transformation layer (STL) to achieve the same goal while removing perturbations as well. Liu et al.~\cite{liu2019feature} proposed a DNN-favorable method called feature distillation based on JPG compression, to rectify adversarial examples without the drop of benign accuracy.

\subsubsection{Proactive Defenses} 
Proactive defenses involve modifying training data or network structure in the training process. Goodfellow et al.~\cite{GoodfellowSS14} injected adversarial examples into the training set and proposed adversarial training, enhancing the robustness of models. Papernot et al.~\cite{papernot2016distillation} introduced defense distillation, which uses the knowledge of network to shape its own robustness. As a result, adversarial examples of small perturbations could be resisted. Xie et al.~\cite{xie2019feature} added novel blocks that denoise features into convolutional networks for robustness improvement.
Other works developed defense from the view of neurons inside the model. Dhillon et al.~\cite{DhillonALBKKA18} developed stochastic activation pruning (SAP) method, which prunes the random subset of the activation function and enlarges remaining ones to compensate. Zhang et al. \cite{zhang2019neuron} traversed all neurons in the model and selected them according to activation values for each class. They designed neuron selecting (NS), achieving defense on small datasets. Suri et al.~\cite{suri2020one} strengthened the robustness of DNNs by targeting neurons that are easily changed by attacks. Bai et al.~\cite{bai2020improving} put forward CAS, which suppresses redundant activation from being activated by adversarial perturbations during training. Zhang et al.~\cite{zhang2020interpreting} explained the adversarial robustness from the perspective of neuron sensitivity, measured by changes of neuron behaviors against benign and adversarial examples. Based on it, they further proposed SNS for retraining the model, finally improving model robustness. Neuron-level defense methods that are relevant to our work are summarized in Table \ref{comparison}. 

In summary, existing works all directly modify neurons. There is no previous work that use neurons’ impact and modify the input value for defense. 

\section{Preliminaries\label{pre}}
\subsection{Threat Model}

The objective of NIP is to defend against attacks without knowing the specific types in advance. For defenders, they can only access part of correctly classified benign examples with their corresponding labels from the training dataset. 

\fm{In the white-box defense setting where the target model is known, the defender uses one layer, such as the flatten layer or global average pooling layer, for defense. In the black-box setting where the structure of the model is unknown, the defender has to train substitute models and use one layer for defense. By default, we use the white-box defense setting in this paper. }

\subsection{\fm{DNNs and Adversarial Attacks}}
A DNN-based classifier can be denoted as $f(x): X\rightarrow Y$, where $X \subset \mathbb{D}^n$ represents the input space and $Y\subset \mathbb{R}^m$ denotes predicted classes. For a given input $x \in X$, $f(x)=\{y_1(x),y_2(x),...,y_m(x)\} \in \mathbb{R}^m$ denotes the confidence values for $m$ class labels. The predicted label $c$ is given as: $c=\text{argmax}_{1\leq i\leq m}y_i(x)$. We denote the corresponding confidence of the predicted label $c$ as $y_c$.

The DNN consists of multiple layers: the input layer, many hidden layers and finally the output layer. Each layer is composed by multiple inner neurons. The output of the layer is the nonlinear combination of each neuron’s activation state in the previous layer. 

The attacker crafts an adversarial perturbation $\Delta x$ for the given benign input $x$, so that the generated adversarial example $x^*=x+\Delta x$ is misclassified by the DNN. Formally, it can be formulated as: $f(x^*) \neq f(x)$. The perturbation $\Delta x$ is limited by $\epsilon$, i.e., $ ||\Delta x|| \leq \epsilon$. $||\cdot||$ represents $l_n$ norm.


\subsection{Definitions}
\quad \textbf{Definition 1} (Neuron's output). Given a DNN and let $X=\{x_1, x_2, ...\}$ denotes a set of inputs. \ff{$N_l=\{n_1, n_2,...\}$ is a set of neurons in the $l$-th layer. $\varphi_n(x)$ represents the output value of neuron $n \in N_l$, when the model is fed with input $x \in X$. The output of $n$ can be calculated as follows:}
\begin{equation}
\varphi_n(x)=\frac{1}{H \times W} \sum_{W} \sum_{H} A_{n}(x) \label{GAP}
\end{equation}
where $A_{n}(x)\in \mathbb{R}^{H\times W\times C}$ is the feature map of $n$ when fed with $x$. $H$ and $W$ denote height and width of the feature map, respectively. 

\textbf{Definition 2} (Neuron influence).  For a neuron $n \in N$ , the neuron influence $\sigma$ of it is defined as:
\begin{equation}
\label{neuron_imp}
    \sigma_{n}=\frac{\partial y_c}{\partial \varphi_n(x)}\cdot \varphi_n(x)
\end{equation}
where $y_c$ denotes the confidence of predicted label $c$. $\partial$ denotes partial derivative function. Neuron influence measures the extent of neuron's contribution to predicted labels: neurons with larger $\sigma$ contribute more to the predicted label.

\textbf{Definition 3} (Front neuron, tail neuron and remaining neuron). According to neuron influence, neurons in the $l$-th layer are sorted in descending order. $top$-$k_1$ and $bottom$-$k_2$ neurons in this neuron sequence are defined as front neurons and tail neurons, denoted as $\Omega_f$ and $\Omega_t$, respectively. Other neurons in this layer are called remaining neurons, denoted as $\Omega_r$. 

\begin{figure}[htbp]
\vspace{-5pt}
\centering
        \includegraphics[width=0.75\linewidth]{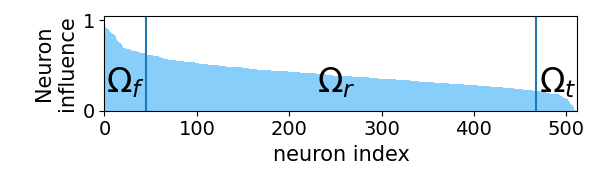} 
        \vspace{-10pt}
\caption{Front neurons $\Omega_f$, remaining neurons $\Omega_r$ and tail neurons $\Omega_t$. Influence values are normalized to [0,1].}
	\label{fig:front}
	\vspace{-10pt}
\end{figure}

\fm{Fig. \ref{fig:front} shows the schematic diagram of front neurons, tail neurons and remaining neurons, where $k_1=k_2$=50.} Flatten layer of VGG19 model on CIFAR-10 dataset is used. 

\section{Approach}
\begin{figure}[t]
\centering
        \includegraphics[width=1\linewidth]{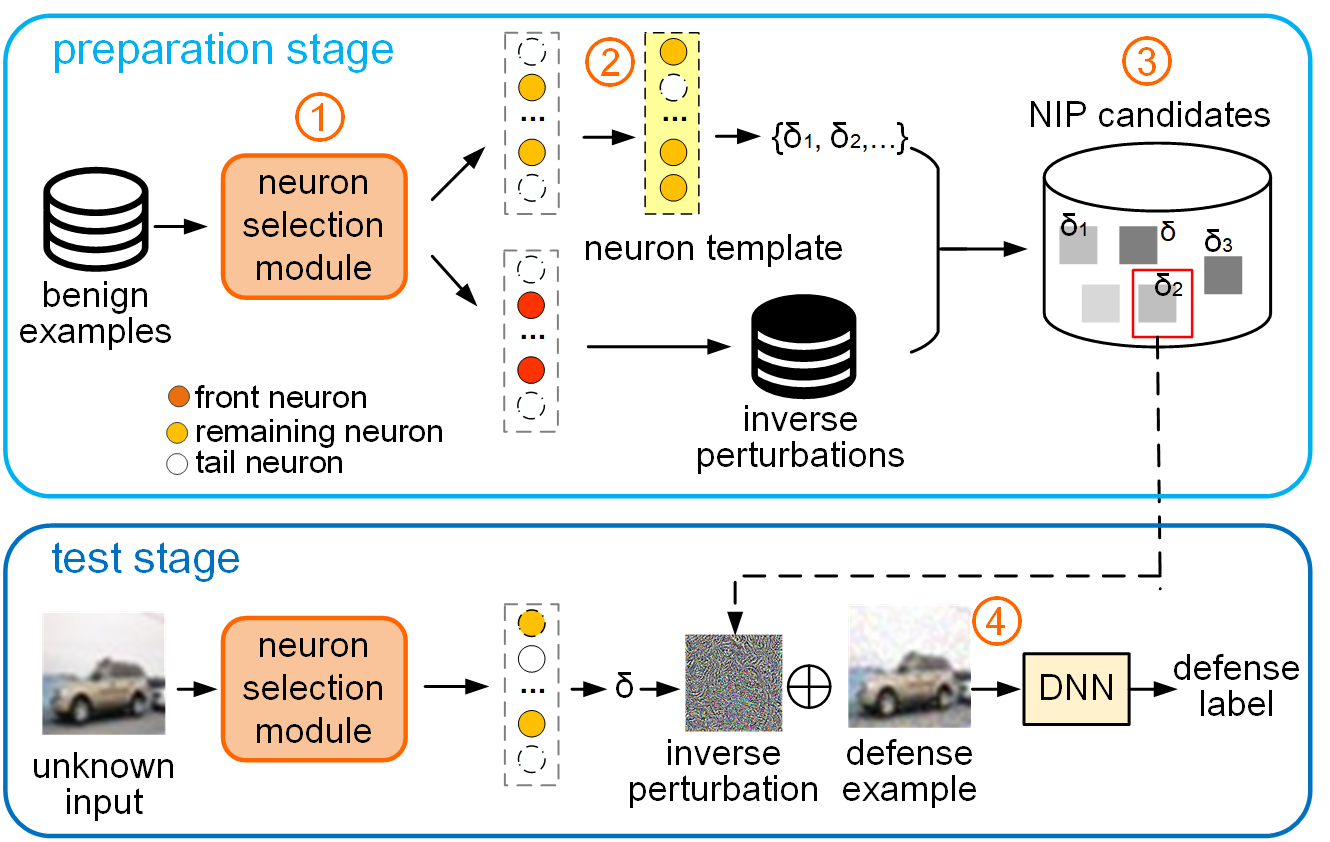} 
        \vspace{-13pt}
\caption{The overview of NIP.}
	\label{fig:framework}
	 \vspace{-5pt}
\end{figure}

\fm{The overview of NIP is shown in Fig.~\ref{fig:framework}, which can be divided into the preparation stage and test stage.} NIP consists of four components: \textcircled{1} neuron selection module, \textcircled{2} generation of neuron template, \textcircled{3} construction of NIP candidates, and \textcircled{4} adaptivity and reclassification. In the test stage, the unknown input can be malicious or benign examples, which is unknown to the defender.

 \begin{algorithm}[htbp]
\caption{Neuron Selection Module}
\label{nsm}
  \renewcommand{\algorithmicrequire}{\textbf{Input:}}
  \renewcommand{\algorithmicensure}{\textbf{Output:}}
\begin{algorithmic}[1]
\REQUIRE Unknown input example $\hat x$ with its predicted label $c$. \ff{Neurons $N_l$ in the $l$-th layer. One neuron $n\in N_l$ with its output function $\varphi_n$.}\\
\ENSURE Sorted neurons $N_l'$.\\
\FOR{$n$ in the $l$-th layer }
		\STATE  Calculate $\sigma_{n}$ according to Equation \ref{neuron_imp}
		\ENDFOR
		\STATE \ff{$N_l'$ $\Leftarrow$ sort $N_l$ in descending order by $\sigma_{n}$}
\end{algorithmic}
\end{algorithm}

In the preparation stage, we use benign examples to conduct \textcircled{1} to \textcircled{3}. Neuron selection module is to select front neurons, remaining neurons and tail neurons in the chosen layer. Remaining neurons are used to generate neuron template, which is further adopted to calculate similarity $\delta$. \fm{Front and tail neurons will be used to generate inverse perturbations, which are paired with their respective $\delta$ to construct NIP candidates.}

\fm{In the test stage, \textcircled{1} and \textcircled{4} are carried out with a given unknown input. The process begins by identifying the remaining neurons to compute the similarity $\delta$ for the given unknown input. Subsequently, according to the $\delta$, an appropriate inverse perturbation is selected from the pool of NIP candidates and applied to the unknown input. This modified input is then fed back to the model for reclassification.} 

In the following, we elaborate on each key component.

\subsection{Neuron Selection Module}
Neuron selection module is adopted to select neurons according to their neuron influence.

\fm{This procedure begins by feeding an unknown input example into the model. We then assess the influence of each neuron within the chosen layer—typically a layer positioned between convolutional and dense layers, such as the flatten layer or global average pooling layer. This choice is motivated by the fact that these layers contain both pixel-level and high-dimensional features crucial for classification tasks.}
Then, front, tail and remaining neurons are selected. $k_1$ and $k_2$ are hyper-parameters for selection. The former $k_1$ is responsible for front neurons while the latter for tail ones. \ff{After that, the output neuron sequence is called sorted neurons, denoted as $N_l'$.} They will be further used for generating neuron template and inverse perturbations. 
The pseudo-code of neuron selection module is presented in Algorithm~\ref{nsm}.

\subsection{Generation of Neuron Template}
\fm{Generated from benign examples, neuron template is a sequence index of remaining neurons.} 

\fm{It is evident from Fig. \ref{intro_attack} that the behavior of remaining neurons remains relatively stable before and after attacks, making them suitable for comparison between benign and unknown inputs. To construct the neuron template, we start with a set of benign examples denoted as $X=\{x_{1},x_{2},...,x_{t}\}$, and calculate neuron template $T$:}
\begin{equation}
    T = \bigcap_{i=1}^{t}\{I(\Omega_{r,x_{i}})\}
\end{equation}
where $\bigcap$ denotes the function of taking intersection. $I(\Omega_{r,x_{i}})$ denotes the index of remaining neurons calculated by the benign example $x_{i}$, and $t$ is the number of benign examples. 

\fm{To ensure compatibility in subsequent calculations, the neuron template $T$ undergoes a specific modification where zero vectors are concatenated to it. This step is crucial for maintaining dimensional consistency with the vector $\Omega_r$. After that, neuron template is marked as $T'$.}

\subsection{Construction of NIP Candidates\label{cons-candidate}}
\fm{The foundational component of NIP is the inverse perturbation, but to accurately match unknown inputs with proper perturbations, we utilize the similarity $\delta$ for accordance. Therefore, each NIP candidate is composed of an inverse perturbation $\rho$ and its corresponding $\delta$, facilitating precise matching of perturbations to the unknown input. By connecting $\rho$ and $\delta$, we can use similarity to find the proper inverse perturbation for the unknown input.}

\fm{For each benign example, we generate one NIP candidate. Specifically, $\rho$ is calculated to enhance the neuron influence of front neurons and suppress that of tail neurons. $\delta$ represents the similarity between the remaining neurons of the input example and those of a predefined neuron template.}

Given a benign example $x \in X$ and template $T'$, the similarity $\delta$ can be calculated by remaining neurons:
\begin{equation}
    \delta_{x}= sim(I(\Omega_{r,x}),T'^{\rm T})
    \label{delta}
\end{equation}
where $sim(\cdot,\cdot)$ measures cosine similarity between two latent vectors. The superscript T means transpose operation.

\fm{In the generation of inverse perturbations, front and tail neurons play crucial roles. Contrary to the typical objectives of adversarial attacks, the optimization goal of NIP involves enhancing the impact of front neurons while suppressing that of tail neurons. This strategy ensures that the inverse perturbations effectively mitigate the adversarial effects. The loss function can be formulated as:}

\begin{equation}
    loss_x= \sum \varphi_{n_j}(x)-\sum \varphi_{n_i}(x)
    \label{loss}
\end{equation}
where $n_i \in \Omega_f$ and $n_j \in \Omega_t$. $\varphi_{n_i}(x)$ denotes the output of the $i$-th neuron in the front neuron $\Omega_{f,x}$. Similarly, $\varphi_{n_j}(x)$ denotes the output of the $j$-th neuron in $\Omega_{t,x}$. $\sum$ denotes the sum function.  
Then the perturbation $\rho_x$ of NIP candidate is generated by taking the partial derivative of the input:
\begin{equation}
    \rho_x=\nabla_x loss_x
    \label{loss_func}
\end{equation}
where $\nabla$ denotes calculating the one-step gradient. 

\fm{Connecting the inverse perturbations with similarity, we can construct NIP candidates, as follows:}
\begin{equation}
    NIP~candidates=\bigcup_{i=1}^{t}\{ \rho_{x_i},\delta_{x_i}\}
\end{equation}
\fm{where $\delta_{x_i}$ and $\rho_{x_i}$ denote $\delta$ and inverse perturbation $\rho$ of $x_i$, respectively. }
The pseudo-code of this process is presented in Algorithm~\ref{cnc}. 
 
\begin{algorithm}[t]
\caption{Construction of NIP Candidates}
\label{cnc}
\begin{algorithmic}[1]
  \renewcommand{\algorithmicrequire}{\textbf{Input:}}
  \renewcommand{\algorithmicensure}{\textbf{Output:}}
\REQUIRE Front neurons $\Omega_f$, remaining neurons $\Omega_r$ and tail neurons $\Omega_t$ of benign example $x \in X $. \\
\ENSURE Inverse perturbation $\rho$ with its similarity $\delta$.\\
\leftline{\textbf{Hyper-parameter:}   $k_1$, $k_2$}
    \STATE Initialize $NIP\ candidates=\{\emptyset\}$
    \FOR{$x \in X$}
    	\STATE $N'$ = Neuron selection module($x$)
        \STATE Calculate $\delta_x$ according to Equation \ref{delta}\\
        \FOR{$i$ in $\Omega_{f,x}$}
            \STATE Front loss $loss_f=\sum \varphi_{n_i}(x)$
            \ENDFOR
        \FOR{$j$ in $\Omega_{t,x}$}
            \STATE Tail loss $loss_t=\sum \varphi_{n_j}(x)$
    	    \ENDFOR
    	\STATE $loss_x=loss_t-loss_f$
    	\STATE Calculate $\rho_x$ according to Equation \ref{loss_func}
    	\STATE $NIP\ candidates\{\cdot\} \Leftarrow \{\rho_x,\delta_x\}$ 
    \ENDFOR
	\RETURN $NIP\ candidates\{\rho, \delta\}$
\end{algorithmic}
\end{algorithm}
	
\subsection{Adaptivity and Reclassification}
\fm{In this process, we identify the appropriate inverse perturbation to apply to the unknown input—such as modifying the pixels in images or the spectral features in audio files. The modified example is then fed back into the model for reclassification. }

\ff{Given an unknown input $\hat{x}$, we can find the suitable inverse perturbation in the candidate pool, according to its similarity $\delta_{\hat{x}}$.}

\fm{The selected inverse perturbation $\rho_{\hat{x}}$ undergoes normalization, then adaptively amplified based on the specific values of each new input. For instance, if the unknown input contains large adversarial perturbations, the inverse perturbation will in turn be amplified to mitigate the adversarial effect.
The defense example $x_{def}$ of $\hat{x}$ is calculated as:}
\begin{equation}
x_{def} = \hat{x} + Mean(\hat{x}) \cdot \frac{\rho_{\hat{x}} }{(Max(\rho_{\hat{x}})-Min(\rho_{\hat{x}}))}
\end{equation}
\ff{where $\rho_{\hat{x}}$ denotes the suitable inverse perturbation of $\hat{x}$ found by its similarity. $Mean(\cdot)$, $Max(\cdot)$ and $Min(\cdot)$ denote the average, maximum and minimum value of the matrix. The inverse perturbation $\rho_{\hat{x}}$ is normalized to ensure consistency in pixel value scale, to align with those of the unknown input. This normalization is crucial as it enhances the imperceptibility of the inverse perturbation. The defense example will be fed into the model for reclassification.}

\textbf{Discussions about the number of NIP candidates.} \fm{The differentiation in neuron templates across various classes necessitates that the number of NIP candidates be at least as extensive as the number of categories. This approach ensures that each category has a tailored inverse perturbation available. However, the practical implementation of this strategy is often hindered by limited access to the full spectrum of benign examples. Defenders typically can utilize only a subset of these examples, which curtails the diversity of NIP candidates that can be generated. Meanwhile, the model must cope with a vast and varied array of unknown inputs.} For measurement, we define $\eta$ as the percentage of the number of NIP candidates to that of unknown inputs: $\eta=\frac{\#~\text{NIP candidates}}{\#~\text{unknown inputs}}$. In our experiment, $\eta$ is set to 1 unless otherwise specified. The impact of it will be discussed in Section \ref{eta}.

\subsection{Algorithm Complexity\label{complexity}}
\fm{We here analyze the complexity of NIP.}

In the preparation stage, we select neurons, generate neuron template and then construct NIP candidates from benign examples. So the computation complexity can be calculated as:
\begin{equation}
    T_{prepare}\sim \mathcal{O}(t \times a)+\mathcal{O}(t \times a) + \mathcal{O}(t \times a)\sim \mathcal{O}(t \times a)
\end{equation}
where $t$ denotes the number of benign examples and $a$ is the number of neurons in the chosen layer.

In the test stage, we select neurons and calculate similarity to find the corresponding NIP. Therefore, the time complexity is:
\begin{equation}
    T_{test}\sim \mathcal{O}(t' \times a)+\mathcal{O}(a) \sim \mathcal{O}(t' \times a)
\end{equation}
where $t'$ denotes the number of unknown inputs.

\section{Experiment Setup\label{setting}}
\subsection{Datasets}

Experiments are conducted on local models on 4 image datasets, including (1) CIFAR-10 and (2) CIFAR-100~\cite{krizhevsky2009learning} - 60,000 32$\times$32 RGB images from 10 and 100 classes, respectively; (3) GTSRB~\cite{Stallkamp-IJCNN-2011} - more than 50,000 German traffic sign images with 48$\times$48 from 43 classes; (4) a-ImageNet - a 10-class subset of over 10,000 animal images with 224$\times$224 from ImageNet~\cite{russakovsky2015imagenet}. For CIFAR-10 and CIFAR-100, 50,000 examples are used for training and 10,000 for validation set. For GTSRB, we split the original training dataset into two parts, i.e., 70\% for training set and 30\% for validation set. For a-ImageNet, 13,000 images are selected for training and 1,300 images for testing. The pixel value of each image is normalized to [0,1]. 


\subsection{Models}
We implement 8 models for image datasets. For CIFAR-10 dataset, VGG19~\cite{simonyan2014very} and AlexNet~\cite{krizhevsky2012imagenet} models are adopted. For CIFAR-100, we train DenseNet \cite{huang2017densely} and SqueezeNet \cite{iandola2016squeezenet}. LeNet-5~\cite{lecun2015lenet} and ResNet20~\cite{he2016deep} are used for GTSRB dataset. For a-ImageNet, VGG19 and MobileNetV1~\cite{howard2017mobilenets} are used. 
Model configurations and classification accuracy of validation set are shown in Table \ref{model_acc}.

\begin{table}[t]
\centering
\caption{Model configurations and accuracy.}
\vspace{-8pt}
\resizebox{0.9\linewidth}{!}{
\begin{tabular}{ccrcrc}
\toprule 
\large
\textbf{Datasets} & \textbf{Models} & \textbf{\#Parameters} & \textbf{\#Layers} & \textbf{\#Neurons} & \textbf{Accuracy} \\ \hline
\multirow{2}{*}{CIFAR-10} & VGG19        & 39,002,738  & 64  & 50,782 & 91.04\% \\
                          & AlexNet      & 87,650,186  & 12  & 18,282 & 91.24\% \\ \hline 		
\multirow{2}{*}{CIFAR-100} & DenseNet        & 824,020  & 349  & 37,258 & 74.61\% \\
                          & SqueezeNet      & 786,088  & 68  & 10,388 & 76.37\% \\ \hline                          
\multirow{2}{*}{GTSRB}    & LeNet-5      & 172,331     & 7   & 291    & 98.26\% \\
                          & ResNet20     & 276,587     & 70  & 2,603  & 98.83\% \\ \hline
\multirow{2}{*}{a-ImageNet} & VGG19        & 139,666,034 & 64  & 50,782 & 95.40\% \\
                          & MobileNetV1  & 3,238,986   & 81  & 33,802 & 98.16\% \\
                          \hline
\multirow{2}{*}{VCTK}     & ResNet34     & 21,315,166  & 158 & 36,254 & 97.27\% \\
                          & Deep Speaker & 24,201,118  & 101 & 28,190 & 98.75\% \\
\bottomrule 
\label{model_acc}  
\end{tabular}}
\vspace{-18pt}
\end{table}

\subsection{Attacks}
We use 13 attacks to generate adversarial examples. White-box attacks include FGSM~\cite{GoodfellowSS14}, BIM~\cite{Kurakin2017Adversarial}, MI-FGSM~\cite{dong2018boosting}, JSMA~\cite{papernot2016limitations}, PGD~\cite{Madry2018Towards}, DeepFool~\cite{moosavi2016deepfool}, UAP~\cite{moosavi2017universal} and EOT~\cite{athalye2018synthesizing}. Black-box attacks contain AUNA~\cite{FoolboxAUNA}, PWA~\cite{SchottRBB19}, One pixel (Pixel)~\cite{su2019one} and Boundary~\cite{Brendel2018Boundary}. Besides, we also conduct AutoAttack (Auto)~\cite{croce2020reliable} for evaluation. Pixel attack is only applied on CIFAR-10 because it fails to fool models for all conducted examples in complex datasets. 

\begin{figure*}[t]
\vspace{-5pt}
\centering
    \subfigure[CIFAR-10, VGG19]{
        \includegraphics[width=0.24\linewidth]{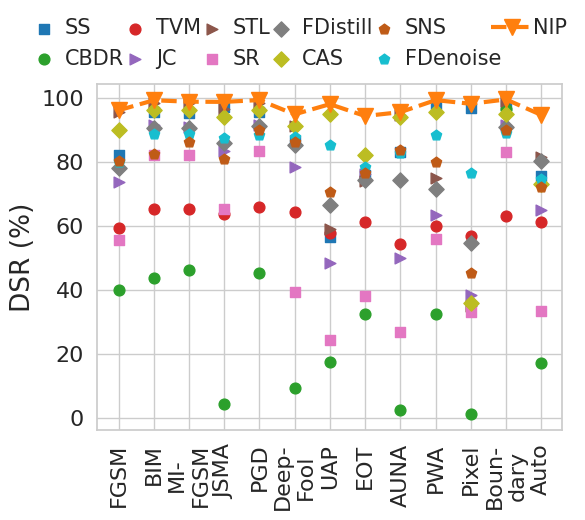}}
    \subfigure[CIFAR-10, AlexNet]{
        \includegraphics[width=0.24\linewidth]{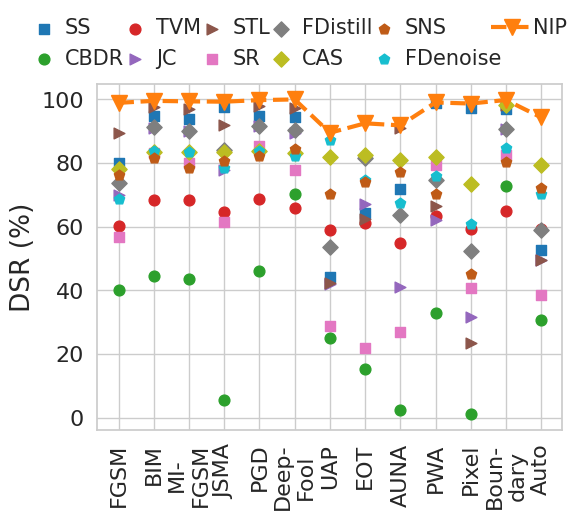}}
    \subfigure[CIFAR-100, DenseNet]{
        \includegraphics[width=0.24\linewidth]{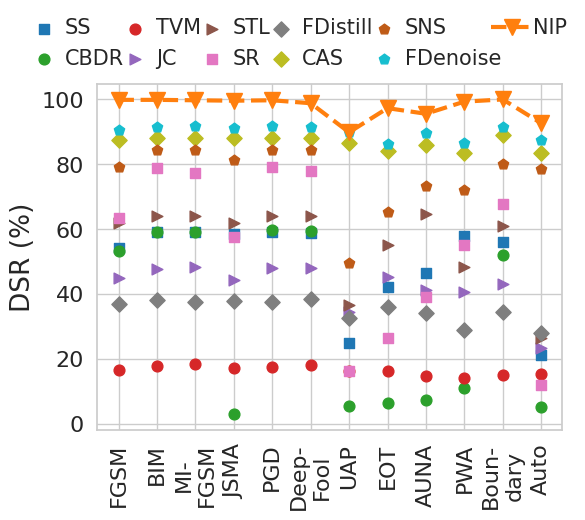}}
    \subfigure[CIFAR-100, SqueezeNet]{
        \includegraphics[width=0.24\linewidth]{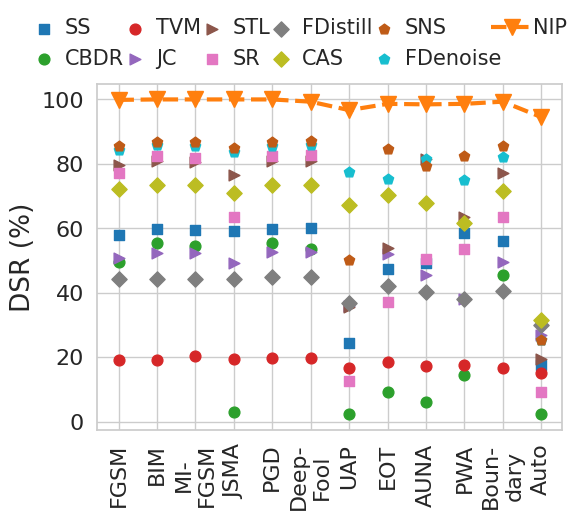}}\\
        \vspace{-8pt}
    \subfigure[GTSRB, LeNet-5]{
        \includegraphics[width=0.24\linewidth]{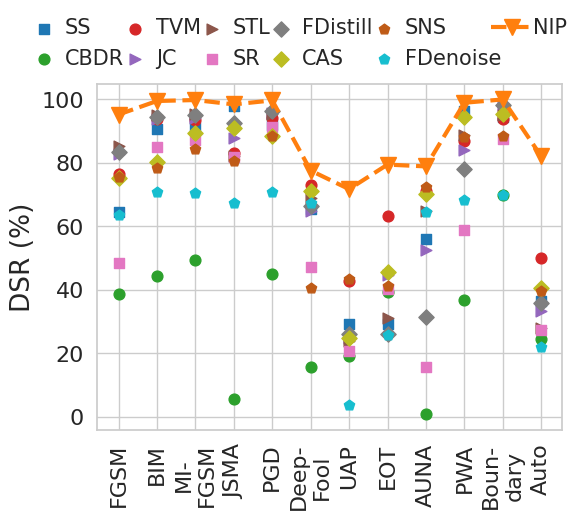}} 
    \subfigure[GTSRB, ResNet20]{
        \includegraphics[width=0.24\linewidth]{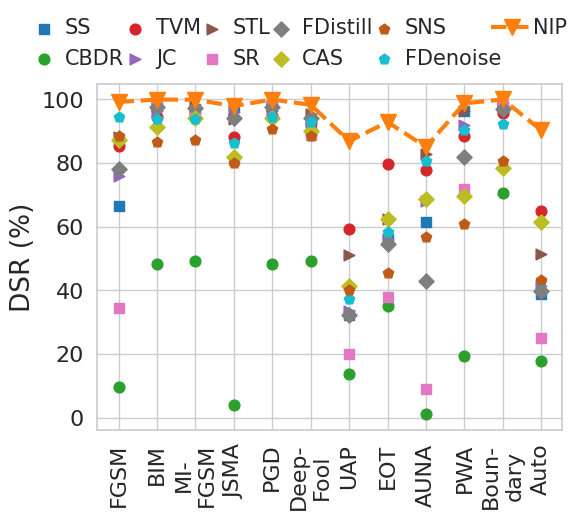}}    
    \subfigure[a-ImageNet, VGG19]{
        \includegraphics[width=0.24\linewidth]{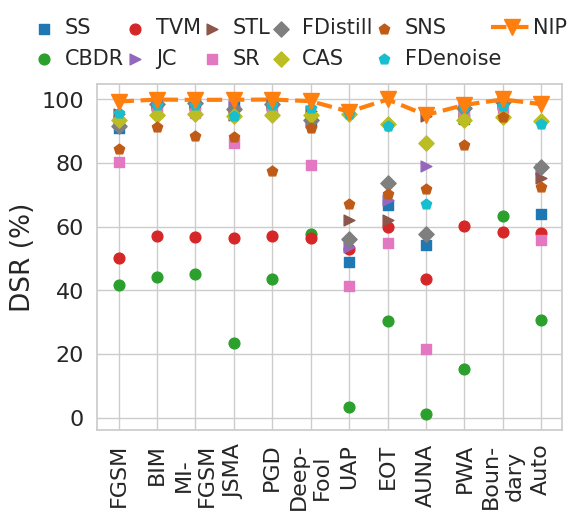}}    
    \subfigure[a-ImageNet, MobileNetV1]{
        \includegraphics[width=0.24\linewidth]{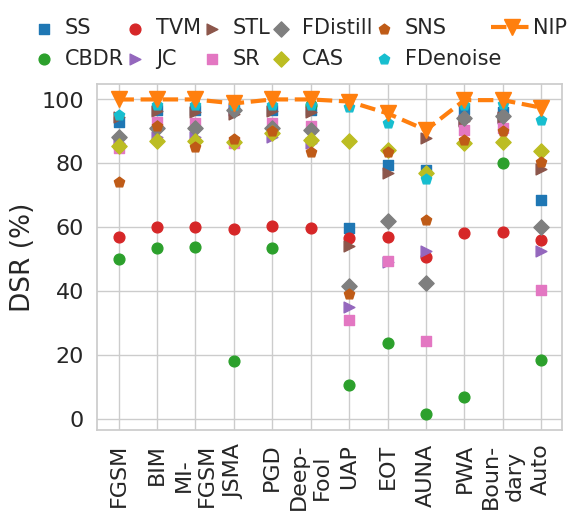}}  \\  
        \vspace{-10pt}
\caption{Comparison of defense results against white-box and black-box attacks on different datasets and models. }
\label{fig:defense}
\vspace{-8pt}
\end{figure*}

\subsection{Baselines}
We implement and compare 10 defense methods with NIP to evaluate their performance. 7 reactive defenses are conducted, including spatial smoothing (SS)~\cite{Xu0Q18}, color bit-depth reduction (CBDR)~\cite{Xu0Q18}, JPG compression (JC)~\cite{dziugaite2016study}, total variance minimization (TVM)~\cite{guo2018countering}, STL~\cite{sun2019adversarial}, super resolution (SR)~\cite{mustafa2019image} and feature distillation (FDistill)~\cite{liu2019feature}. Besides, 3 proactive defenses, CAS~\cite{bai2020improving}, SNS~\cite{zhang2020interpreting} and feature denoising (FDenoise)~\cite{xie2019feature}, are also adopted as baselines. They are configured according to the best performance setting reported in the respective papers. We implement our copy of SNS following the introduced technical approach.


\subsection{Evaluation Metrics} 
To evaluate defense performance, we use 3 metrics, as follows: 

(1) classification accuracy (acc): $acc=\frac{N_{true}}{N_{benign}}$, where $N_{true}$ is the number of benign examples correctly classified by the targeted model and $N_{benign}$ denotes the total number of benign examples.


(2) attack success rate (ASR): $ASR = \frac{N_{adv}}{N_{benign}}$, where $N_{adv}$ denotes the number of adversarial examples.

(3) defense success rate (DSR): $DSR = \frac{N_{def}}{N_{adv}}$, where $N_{def}$ denotes the number of adversarial examples correctly classified by the targeted model after defense. Intuitively, higher defense success rate indicates more effective defenses.


\subsection{Implementation Details}
Our experiments have the following settings: (1) We set $k_1$ and $k_2$=5, $\eta$=1 in NIP, unless otherwise specified. Global average pooling is chosen for DenseNet, SqueezeNet and MobileNetV1, while flatten layer is chosen for other models. (2) To mitigate non-determinism, we repeat the experiment for 3 times and reported the average results. (3) More detail settings about adversarial attacks and defense baselines are shown in the \textbf{appendix}.

All experiments are conducted on a server under Ubuntu 20.04 operating system with Intel Xeon Gold 5218R CPU running at 2.10GHz, 64 GB DDR4 memory, 4TB HDD and one GeForce RTX 3090 GPU card.

\section{Experimental Results}
We evaluate the performance of NIP by answering the following six research questions (RQs):  
\begin{itemize} [nolistsep, leftmargin=*]
     \item \textbf{RQ1: Effectiveness} - Is NIP more effective against various attacks, when compared with SOTA baselines? 
     \item \textbf{RQ2: Elasticity} - Is NIP robust to different perturbation sizes? 
      \item \textbf{RQ3: Efficiency} - \ff{How efficient is NIP in defense? }
      \item \textbf{RQ4: Extensibility} - \ff{Can NIP be applied to speaker recognition systems and online image platforms? }
\end{itemize}

\subsection{RQ1: Effectiveness}
When reporting the results, we focus on the following aspects: defense against various attacks and its impact on benign examples. We have provided comparisons of effectiveness between neuron-level defenses and NIP in the \textbf{appendix}.

\begin{table}[t]
\centering
\huge
\caption{The p-value of t-test between NIP and each baseline. }
\vspace{-8pt}
\resizebox{1\linewidth}{!}{
\begin{tabular}{cccccccccccc}
\toprule
\multirow{2}{*}{\textbf{Datasets}} &
  \multirow{2}{*}{\textbf{Models}} &
  \multicolumn{10}{c}{\textbf{Defenses}} \\ \cline{3-12} 
 &    &   SS & 
  CBDR &  TVM &  JC &  STL &  SR &  FDistill &  CAS &  FDenoise &  SNS \\ \hline
\multirow{2}{*}{CIFAR-10} & VGG19 & 0.0085   & 0.0000 & 0.0000  & 0.0003 & 0.0310 & 0.0000  & 0.0000      & 0.0468 & 0.0001  & 0.0001 \\  	 	 	 	 	 	 
 &
  AlexNet & 0.0070   & 0.0000   & 0.0000  & 0.0003  & 0.0138  & 0.0000  & 0.0000  &0.0000   & 0.0002  & 0.0000  \\ \hline
 \multirow{2}{*}{CIFAR-100} & DenseNet & 0.0000     & 0.0000 & 0.0000 & 0.0000 & 0.0000 & 0.0001 & 0.0000   & 0.0000 & 0.0000 & 0.0000 \\
 &  	 	 	 	 	 	 	 	 	 
  SqueezeNet &  0.0000  & 0.0000   & 0.0000  & 0.0000  & 0.0002  & 0.0005  & 0.0002  & 0.0000  & 0.0000  &  0.0013 \\ \hline	 
\multirow{2}{*}{GTSRB} &  LeNet-5 & 0.0035  & 0.0000  & 0.0002  & 0.0019  & 0.0015  & 0.0009  & 0.0131  &  0.0007 & 0.0000  &   0.0003  \\ 
 & ResNet20 & 0.0484 & 0.0000  & 0.0021  & 0.0439  & \textbf{0.0818}  & 0.0205  & 0.0310  & 0.0021  & 0.0336 & 0.0007   \\ \hline  	 	 	 	 	 	 	 	
\multirow{2}{*}{a-ImageNet} & VGG19 & 0.0159 & 0.0000 & 0.0000 & 0.0202 & 0.0292  & 0.0393 & 0.0058 & 0.0000  & 0.0167 & 0.0000 \\ 									
 &  MobileNetV1 & 0.0129  & 0.0000  & 0.0000  & 0.0028  & 0.0171  & 0.0176  & 0.0042  & 0.0000 & 0.0041  &  0.0009 \\ \bottomrule
\label{pvalue}
\end{tabular}}
\vspace{-20pt} 
\end{table}

\subsubsection{Defense Results against Various Attacks\label{defense}}
We focus on the NIP's defense effectiveness against various adversarial examples. 

\begin{table*}[htbp]
\centering
\caption{Comparison of benign accuracy after defense.}
\vspace{-8pt}
\resizebox{0.7\linewidth}{!}{
\large
\begin{tabular}{ccccccccccccc}
\toprule
\multirow{2}{*}{\textbf{Datasets}} &
  \multirow{2}{*}{\textbf{Models}} &
  \multicolumn{11}{c}{\textbf{Defenses}} \\ \cline{3-13} 
 &
   &
  SS &
  CBDR &
  TVM &
  JC &
  STL &
  SR &
  FDistill &
  CAS &
  FDenoise &
  SNS &
  NIP \\ \hline
\multirow{2}{*}{CIFAR-10} &
  VGG19 &
  90.80\% &
  90.95\% &
  62.91\% &
  89.04\% &
  90.76\% &
  76.84\% &
  90.31\% &
  84.30\% &
  80.57\% &
  82.30\% &
  \textbf{92.51\%} \\
 &
  AlexNet &
  90.45\% &
  90.15\% &
  65.33\% &
  88.23\% &
  91.47\% &
  90.56\% &
  89.19\% &
  85.77\% &
  85.40\% &
  84.03\% &
  \textbf{92.59\%} \\ \hline
\multirow{2}{*}{CIFAR-100} & DenseNet & 58.11\% & 74.61\% & 15.15\% & 45.65\% & 72.61\% & 66.49\% & 31.71\% & 65.24\% & 68.68\% & 64.09\% & \textbf{75.01\%} \\
 &
  SqueezeNet &
  61.71\% &
  \textbf{76.25\%} &
  16.72\% &
  52.84\% &
  74.80\% &
  68.34\% &
  39.33\% &
  55.99\% &
  65.38\% &
  67.43\% &
  76.20\% \\ \hline
\multirow{2}{*}{GTSRB} &
  LeNet-5 &
  \textbf{97.57\%} &
  97.47\% &
  96.98\% &
  97.47\% &
  97.28\% &
  97.38\% &
  97.57\% &
  94.53\% &
  94.03\% &
  95.61\% &
  95.51\% \\
 &
  ResNet20 &
  98.43\% &
  98.73\% &
  98.43\% &
  98.63\% &
  97.84\% &
  98.58\% &
  98.14\% &
  93.30\% &
  95.17\% &
  96.85\% &
  \textbf{98.93\%} \\ \hline
\multirow{2}{*}{a-ImageNet} &
  VGG19 &
  89.49\% &
  91.01\% &
  54.19\% &
  88.91\% &
  90.63\% &
  \textbf{94.92\%} &
  89.39\% &
  91.01\% &
  91.30\% &
  89.29\% &
  94.54\% \\
 &
  MobileNetV1 &
  94.14\% &
  94.92\% &
  78.04\% &
  95.22\% &
  95.57\% &
  74.60\% &
  95.31\% &
  93.55\% &
  96.98\% &
  91.49\% &
  \textbf{98.61\%} \\\bottomrule
 \label{benign}
\end{tabular}}
 \vspace{-13pt}
\end{table*}

\textbf{Implementation Details.} We adopt 13 adversarial attacks and use 5,000 adversarial examples per attack to calculate DSR on 8 image classification models after defense. Results of NIP and baselines are shown in Fig. \ref{fig:defense}. The p-value of t-test between NIP and each baseline is also calculated using two-tail and paired-sample t-test. Results are reported in Table \ref{pvalue}, where p-value larger than 0.05 is bold. 

\textbf{Results and Analysis.} In all cases, NIP significantly improves the model's classification towards adversarial attacks. It outperforms both proactive and reactive defense baselines with a considerable margin. In Fig. \ref{fig:defense}, almost all DSR of NIP are the highest. 
And almost all p-values of NIP in Table \ref{pvalue} are smaller than 0.05. This is mainly because it can alleviate and even offset the effect of adversarial perturbations, by strengthening front neurons and weakening tail ones that exploited by attacks.

NIP shows more stable defense performance among various adversarial attacks, regardless of attack types or image scales. 
For instance, on AlexNet of CIFAR-10, DSR of NIP is around 98\% on average against various attacks, while DSR of SNS shows sharp decrease on PWA. We speculate the possible reason is that NIP implements defense based on general observations on neuron behaviors, so it can deal with perturbations from various attacks. 

With regard to multi-class dataset like GTSRB and CIFAR-100, the effectiveness of NIP is also guaranteed on a majority of attacks. This is because NIP links adversarial examples with corresponding inverse perturbation. That is more fine-grained solution, since we nearly generate NIP tailored to each input image.
Due to the shift of the feature during mapping or projection process, STL and SR do weaken the attack strength, but cannot achieve better defense among all attacks. CAS and SNS, as adversarial training strategies, show inferior performances. Based only on PGD, SNS reduces neuron sensitivity during retraining while both activation values and frequency are suppressed by CAS. These methods help construct robust models but fail to handle other perturbations that not included during training.

\subsubsection{Defense Impact on Benign Examples}
Since some existing defenses are compromised to the accuracy of benign examples, in this part we focus on the impact of NIP on benign examples.

\textbf{Implementation Details.} For each model, we feed benign examples and count the classification accuracy after defense. Results are shown in Table \ref{benign}. Higher accuracy indicates less negative impact on benign examples. 

\textbf{Results and Analysis.} From Table \ref{benign}, NIP achieves highest benign accuracy in most cases. This indicates that NIP hardly sacrifices the benign accuracy while completing effective defense. Some increase on benign accuracy of NIP can also be observed, e.g., 92.59\% on AlexNet of CIFAR-10 and 98.61\% on MobileNetV1 of a-ImageNet. We think misclassifications of some benign examples are mainly due to the equal importance of neurons. NIP enhances class-related front neurons, which helps to correct labels of misclassified benign examples. AS for baselines, TVM and SR remove adversarial perturbations together with important pixels, which decreases the accuracy on benign examples after defense. CAS and SNS show inferior results on benign accuracy, especially on CIFAR-10 and CIFAR-100. This is mainly because they suppress some class-relevant neurons repeatedly during training.

\begin{center}
\fcolorbox{black}{gray!20}{\parbox{0.95\linewidth}
    {
        \textbf{Answer to RQ1:} NIP outperforms the SOTA baselines in effectiveness in two aspects: (1) DSR against various attacks ($\sim \times 1.45$ on average); (2) higher benign accuracy after defense. 
    }
}
\vspace{-5pt}
\end{center}

\subsection{RQ2: Elasticity}
When reporting the results, we focus on defense elasticity on different perturbation sizes. 

\begin{figure}[htbp]
\vspace{-15pt}
\centering
    \subfigure[CIFAR-10, VGG19]{
        \includegraphics[width=0.45\linewidth]{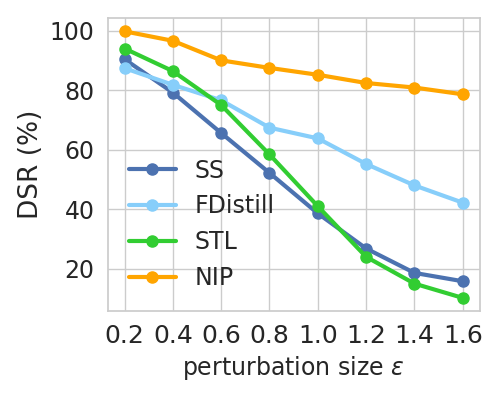}}
    \subfigure[a-ImageNet, MobileNetV1]{
        \includegraphics[width=0.50\linewidth]{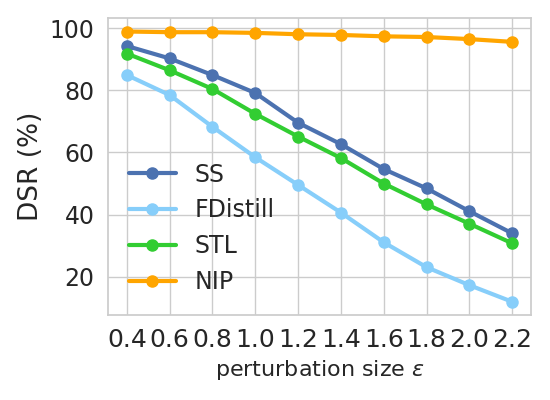}}
        \vspace{-10pt}
\caption{DSR of NIP and three baselines under different perturbation sizes.}
\label{bigp}
\vspace{-8PT}
\end{figure}

\textbf{Implementation Details.} We conduct experiments on VGG19 of CIFAR-10 and MobileNetV1 of a-ImageNet. 1,000 adversarial examples generated by PGD are used. $l_2$-norm perturbation size $\epsilon$ from 0.2 to 1.6 added on CIFAR-10, while $\epsilon$ from 0.4 to 2.2 for a-ImageNet. We compare NIP with SS, FDistill and STL due to their superior defense effect, according to Section \ref{defense}. We calculate DSR per $\epsilon$ and report them in Fig. \ref{bigp}.

\textbf{Results and Analysis.} NIP shows stronger robustness on defense effect than baselines, among all perturbations. For instance, orange lines are the highest in Fig. \ref{bigp}. With the increase of perturbation size, slight drop of DSR of NIP can be observed. Specifically, when $\epsilon$ reaches 1.6, baselines (i.e., STL and SS) can hardly take effect on CIFAR-10, with DSR around 30\%. But DSR of NIP still exceeds 70\%. It is mainly because adversarial perturbations of different sizes can be alleviated and removed by NIP via adaptively amplifying the inverse perturbations sizes. This well demonstrate the effectiveness of adaptivity before reclassification. However, defense based on feature distribution are compromised in the event of large perturbations, let alone simple image transformations. They fail to restore the feature distribution of benign examples, leading to unsatisfactory results.

\begin{center}
\fcolorbox{black}{gray!20}{\parbox{0.95\linewidth}
    {
        \textbf{Answer to RQ2:} NIP shows better elasticity than SOTA baselines on perturbations of different sizes: It shows quite stable DSR ($\sim \times 3.4$) when $l_2$-norm perturbation size is larger than 1.4.
    }
}
\vspace{-5pt}
\end{center}

\subsection{RQ3: Efficiency}
In this part, we give a comparison of defense time between NIP and reactive baselines.

\textbf{Implementation Details.} The average defense time of NIP and reactive baselines are calculated in the test stage. We measure average running time on 5,000 adversarial examples crafted by each attack for 3 times, and the minimal one is identified as the final result, as shown in Table \ref{time}. 

\begin{table}[t]
\centering
\caption{Comparison of time complexity between NIP and reactive baselines in the test stage. }
\vspace{-8pt}
\Huge
\resizebox{1\linewidth}{!}{
\begin{tabular}{cccccccccc}
\toprule
\multirow{2}{*}{\textbf{Datasets}}  & \multirow{2}{*}{\textbf{Models}} & \multicolumn{8}{c}{\textbf{Defenses}}                                                \\ \cline{3-10} 
                          &                        & SS    & CBDR  & TVM      & JC    & STL       & SR     & FDistill & NIP     \\ \hline
\multirow{2}{*}{CIFAR-10} & VGG19                  & 3.13  & 2.11  & 219.49   & 3.96  & 26716.16  & 138.57 & 142.84   & 195.25 \\
                          & AlexNet                & 2.47  & 1.43  & 194.03   & 2.89  & 14559.11  & 282.40 & 63.38    & 225.48 \\ \hline
\multirow{2}{*}{CIFAR-100} & DenseNet                  & 3.34  & 2.77  & 211.00   & 4.00  &  4326.74 & 133.58 & 46.18   & 4704.02 \\  	 	 			 	 	 
                          & SqueezeNet                & 2.45  &  1.87 & 215.24   & 2.98  & 4156.39  & 91.00 &  44.75   & 1071.50  \\ \hline	
\multirow{2}{*}{GTSRB}    & LeNet-5                & 4.15  & 1.25  & 458.59   & 2.75  & 9256.90   & 273.15 & 112.25   & 110.56 \\
                          & ResNet20               & 5.81  & 2.22  & 440.54   & 3.84  & 10061.09  & 232.49 & 123.58   & 303.56 \\ \hline
\multirow{2}{*}{a-ImageNet} & VGG19                  & 68.99 & 33.48 & 15030.03 & 42.07 & 115870.37 & 641.13 & 2046.99  & 345.85 \\
                          & MobileNetV1            & 74.51 & 44.08 & 8352.64  & 37.87 & 124519.79 & 720.77 & 2875.80  & 305.43 \\ \bottomrule
\label{time}
\end{tabular}
}
\vspace{-20pt}
\end{table}

\textbf{Results and Analysis.} From Table \ref{time}, running time of NIP is acceptable, especially on large dataset like a-ImageNet. For instance, time of NIP on CIFAR-10 is around 200s, inferior to that of SR and FDistill slightly. But on a-ImageNet, NIP requires much less time (around 300s) than TVM (around 1000s), STL (over 10000s) and FDistill (over 2000s). In the testing stage, NIP searches for the closest inverse perturbation. So the computation time of it is only related to the number of examples under defense. For baselines like STL and FDistill, more computation time is required on large images for they contains more pixels. But compared with simple image transformations like SS, CBDR and JC, NIP shows disadvantage, especially on complex models like DenseNet. This is mainly because it involves the model structure and traverses all neurons in the selected layer to calculate neuron influence. 

\begin{center}
\fcolorbox{black}{gray!20}{\parbox{0.95\linewidth}
    {
        \textbf{Answer to RQ3:} NIP is more efficient in defense: In the test stage, NIP achieves defense using 1/6 computation time on average, compared with reactive baselines. 
    }
}
\vspace{-5pt}
\end{center}

\subsection{RQ4: Extensibility}
Apart from image classification on local models, NIP can be extended to different classification tasks. We evaluate the performance of it on speaker recognition local models and online platforms.

\subsubsection{On Speaker Recognition Models}
Speaker recognition systems are widely used in the real-world scenarios. In this part, we demonstrate the application of NIP in speaker recognition models.

\begin{figure}[t]
\vspace{-4pt}
\centering
    \subfigure[ResNet34]{
        \includegraphics[width=0.48\linewidth]{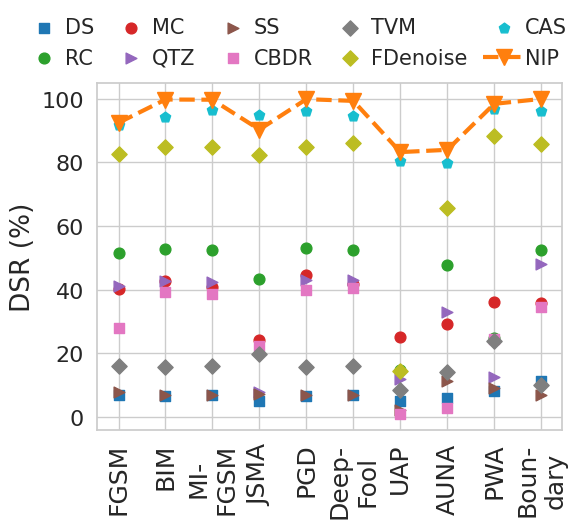}}
    \subfigure[Deep Speaker]{
        \includegraphics[width=0.48\linewidth]{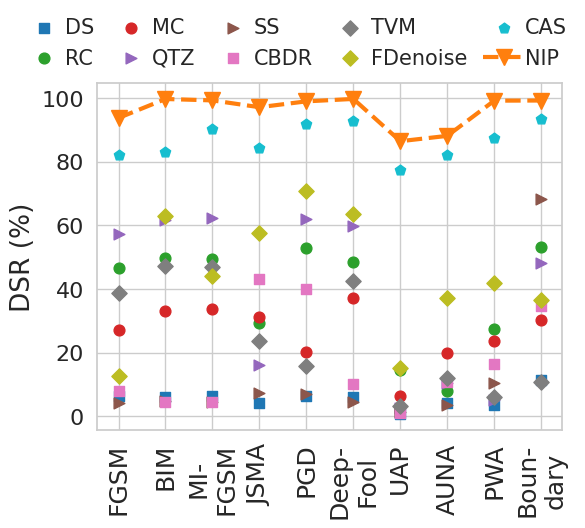}}
        \vspace{-10pt}
\caption{Comparison of defense results on VCTK datasets.}
\label{fig:defense_audio}
\vspace{-10pt}
\end{figure}

\textbf{Implementation Details.} NIP is evaluated on ResNet34~\cite{he2016deep} and Deep Speaker~\cite{li2017deep} on VCTK dataset~\cite{yamagishi2019cstr}, which consists of more than 10,000 audios of native English speakers from 30 classes. 10,000 examples from  are used for training, and 1,200 are for validation. \fm{Both models leverage CNNs, similar to those used in image processing, to capture spatial hierarchies in the audio data.} Model configurations are also shown in Table \ref{model_acc}. We generate 4,000 adversarial examples with same attacks as Section~\ref{setting} for each model. We calculate DSR and show results in Fig. \ref{fig:defense_audio}. we adopt 9 defense baselines, including 4 pre-processing methods designed specifically for audio data: Down sampling (DS)~\cite{YangLCS19}, Random crop (RC)~\cite{li2006localized}, MP3 compression (MC)~\cite{andronic2020mp3} and Quantization (QTZ)~\cite{YangLCS19}. Besides, SS, CBDR, TVM, CAS and FDenoise are still adopted for fair comparison. 

\textbf{Results and Analysis.} It is obvious that NIP achieves higher DSR (95\% on average) on VCTK dataset, outperforming other defense baseline, regardless of audio-specific or image-specific defense methods. \fm{This also supports and extends our observation that adversarial audio examples often exploit different neurons than those by benign examples. NIP targets those front and tail neurons, to generate inverse perturbations that can mitigate the adversarial effect.} DS, RC, MC and QTZ, which are designed specifically for audios, show inferior and unstable performance (i.e., about 50\%), especially on large perturbations. Therefore, we believe NIP is helpful in defending adversarial attacks and building more robust intelligent systems. 

\subsubsection{On Online Platforms}
We apply NIP to Baidu online platforms for animal\footnote{Visit \emph{https://cloud.baidu.com/product/imagerecognition/animal} for animal recognition.} and plant\footnote{Visit \emph{https://cloud.baidu.com/product/imagerecognition/plant} for plant recognition.} recognition, to further verify the effectiveness of NIP. 

\textbf{Implementation Details.} We train two self-developed VGG19 models as local substitute models. For animal recognition, a-ImageNet is used while 17 Category Flower Dataset, containing 1360 images belonging to 17 classes, are used for plant recognition. AUNA attack with high transferability is adopted to craft adversarial examples for online platforms, with ASR 100\% for two self-trained substitute models. We evaluate NIP on 500 adversarial examples and calculate DSR after defense. 
If the $top$-1 prediction tag gained from online platform response is different from the original natural one, it is considered as a successful attack. Similarly, if the response of examples after defense returns same as the original one, the defense is successful. For measurement, we call API provided by Baidu and calculated DSR according to the response.

\textbf{Results and Analysis.} The DSR of NIP on animal and plant recognition is 97.20\% and 93.00\%, respectively. Our experimental results demonstrate that NIP is effective when applied to online image classification platforms. We can also conclude that in real-world scenarios that models remain black-box to defenders, NIP can still take effect by training local substitute models. 

\begin{center}
\fcolorbox{black}{gray!20}{\parbox{0.95\linewidth}
    {
        \textbf{Answer to RQ4:} Apart from self-developed image classification models, NIP shows defense extensibility in (1) speaker recognition models and (2) Baidu online image platforms. 
    }
}
\vspace{-5pt}
\end{center}

\section{Discussions}
In this section, we discuss adaptive attacks, parameter sensitivity and provide interpretable analysis for defense. More results, visualizations and correlation between neuron template and labels can be found in the \textbf{appendix}.

\subsection{Potential Countermeasures}
It must be noted that if the defense behavior is deterministic for the same adversarial example, the attacker can adaptively adopt specific methods against defense. To assess the feasibility of this idea, we design two types of adaptive attacks: smooth adaptive loss attack (S-ALA) and remaining neuron adaptive loss attack (R-ALA).

(1) S-ALA. It achieves attacks while maintaining similar patterns of neuron influence with benign examples. Adversarial examples are calculated by:
\begin{equation}
    \mathop{\arg\min}\limits_{x^*}~ MSE(\bigcup_{i=1}^{a}\sigma_{n_i}(x), \bigcup_{i=1}^{a}\sigma_{n_i}(x^*))- \lambda\mathcal{J}(\theta,x^*,y)
\end{equation}
where $MSE(\cdot, \cdot)$ is the mean square error and $\lambda$ is the balancing parameter, which is set to 1 by default. $\mathcal{J}$ is the cross-entropy loss of the model with parameters $\theta$, adversarial example $x^*$ and truth label $y$. $n\in N$ is a single neuron in the chosen layer and the total number of neurons in $N$ is $a$. We calculate neuron influence of each neuron in flatten layer or GAP for $\sigma(x)$. The calculation of each adversarial example is iterated repeatedly for 10 rounds. 

(2) R-ALA. \fm{It fools the model while trying to larger the difference of the remaining neurons with benign examples.} Adversarial examples are calculated by:
\begin{equation}
    \mathop{\arg\max}\limits_{x^*}~ MSE(\bigcup_{j=1}^{b}\sigma_{n_j}(x), \bigcup_{j=1}^{b}\sigma_{{n_j}}(x^*))+ \lambda\mathcal{J}(\theta,x^*,y)
\end{equation}
where $n\in \Omega_{r}$ is a single neuron in the set of remaining neurons, which contains $b$ neurons in total. We use MSE to control the remaining neurons on flatten layer or GAP. $\lambda$ is also set to 1 here. The calculation is iterated for 10 rounds.

\textbf{Implementation Details.} Experiments are conducted on AlexNet of CIFAR-10, ResNet20 of GTSRB and VGG19 of a-ImageNet. We set $k_1$=$k_2$=5 for NIP. We calculate ASR of these adaptive attacks on 1,000 adversarial examples, as shown in Table \ref{adaptive_asr}. More results of S-ALA on different $\lambda$ and iterations are shown in the \textbf{appendix}.

\textbf{Results and Analysis.} We can observe from Table \ref{adaptive_asr} that although both adaptive attacks are theoretically able to bypass the proposed defense, they are not effective against NIP in practice. 

S-ALA shows strong attack effect, with ASR around 25\%. By matching the internal of adversarial examples with that of benign ones, neuron influence doesn't show large difference. But in that case, front and tail neurons are still responsible for correct and erroneous classification, respectively. This is still consistent with our assumption. \fm{Therefore, by enhancing front neurons and suppressing tail ones, most adversarial examples can be successfully defended.}

Besides, although remaining neurons of adversarial examples are quite different to those of benign ones, NIP can still reach average DSR over 70\% against R-ALA. This shows the generality of neuron template and neuron candidates. Front and tail neurons play important roles in defense. Hence, by controlling them to generate inverse perturbations, we can mitigate the effect of adversarial examples.

\textbf{\begin{table}[t]
\huge
\centering
\vspace{-5PT}
\caption{ASR of adaptive attacks.}
\vspace{-8pt}
\resizebox{0.5\linewidth}{!}{
\begin{tabular}{cccc}
\toprule
\multirow{2}{*}{\textbf{Datasets}} & \multirow{2}{*}{\textbf{Models}} & \multicolumn{2}{c}{\textbf{Attacks}} \\ \cline{3-4} 
                    &          & \textbf{S-ALA} & \textbf{R-ALA} \\ \hline
CIFAR-10   & AlexNet  & 31.10\%        & 25.60\%        \\
GTSRB      & ResNet20 & 21.90\%        & 19.40\%        \\
a-ImageNet & VGG19    & 26.40\%        & 14.40\%        \\ \bottomrule
\label{adaptive_asr}
\end{tabular}}
\vspace{-22pt}
\end{table}}

\subsection{Parameter Sensitivity Analysis}
In this part, we study the impact of different hyper-parameters on the defense effect of NIP, including $k$, chosen layer and candidate ratio $\eta$.

\subsubsection{Influence of $k$} 
\begin{figure}[t]
\centering
    \subfigure[Influence of $k_1$, when $k_2=5$]{
        \includegraphics[width=0.46\linewidth]{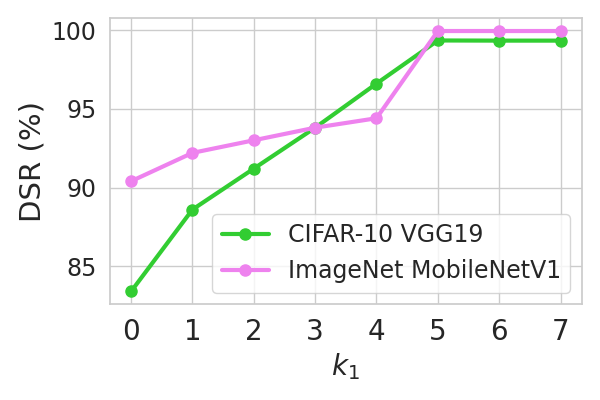}}
    \subfigure[Influence of $k_2$, when $k_1=5$]{
        \includegraphics[width=0.46\linewidth]{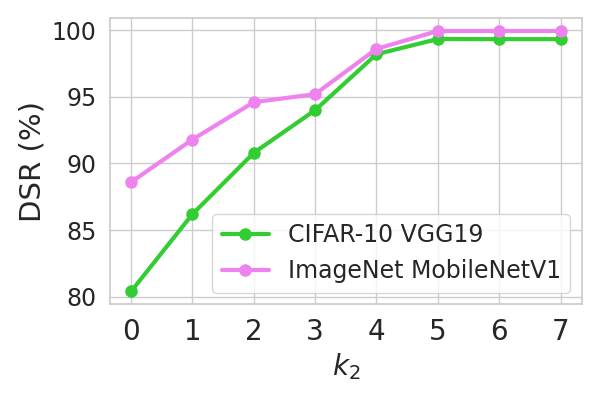}}
        \vspace{-12pt}
\caption{Results of DSR under different $k_1$ and $k_2$.}
\label{k1k2}
\vspace{-15pt}
\end{figure}

$k_1$ and $k_2$ are two hyper-parameters responsible for selecting front and tail neurons, respectively. We calculate DSR on 1,000 adversarial examples crafted by PGD using different $k_1$ and $k_2$. Results are shown in Fig. \ref{k1k2}.

DSR shows upward trend with the increase of $k_1$ and $k_2$. When $k_1$ and $k_2$ are larger than 5, DSR remains stable (around 100\%). Controlling more front and tail neurons will lead to enhancement of more class-relevant pixels and offset more perturbed pixels. Hence, DSR increases in this process. When $k_1$ and $k_2$ are set larger than 5, neurons that play the most decisive roles on correct classifications are nearly taken into consideration, so DSR remains stable. 

It must be noted that NIP still shows considerable DSR (over 80\%) when $k_1$ or $k_2$ is either set to 0. Concretely, either strengthening front or suppressing tail neurons can rectify classifications. Furthermore, the modification of $k_2$ has a more significant impact on DSR than $k_1$. For instance, on MobileNetV1, DSR is over 90\% when $k_1$=0, higher than 88\% when $k_2 =0$. It indicates that suppressing tail neurons that may be utilized by attacks works more effectively. But better defense performance can be guaranteed with the blend of controlling both front and tail neurons. 

\begin{figure}[t]
\centering
    \subfigure[CIFAR-10, VGG19]{
        \includegraphics[width=0.47\linewidth]{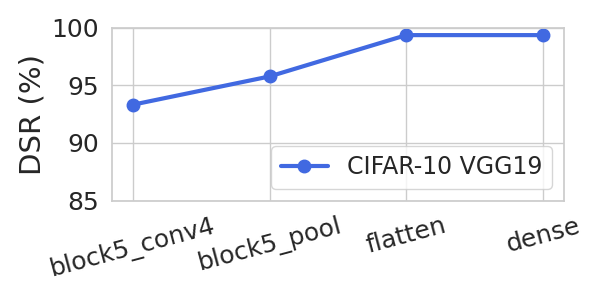}}
    \subfigure[ImageNet, MobileNetV1]{
        \includegraphics[width=0.47\linewidth]{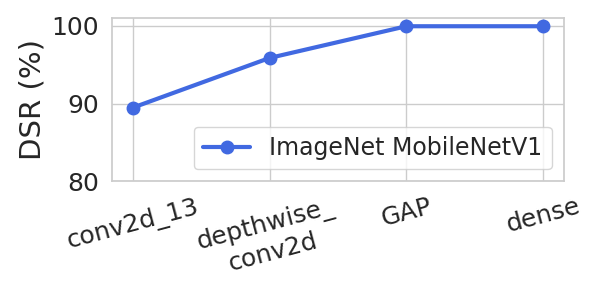}}
\caption{The results of DSR against PGD under different chosen layer on VGG19 of CIFAR-10 (left) and MobileNetV1 of a-ImageNet (right).}
\label{layer}
\vspace{-16pt}
\end{figure}

\subsubsection{Influence of Network Layer}
The strategy of choosing layer can be tricky, so we change the layer of each targeted model to study the influence of chosen layer. The experimental results against PGD are shown in Fig. \ref{layer}, 
where the x-axis denotes the name of chosen layer in each model. 
``GAP'' is short for global average pooling.

As observed, DSR increases as chosen layer goes deeper. 
It reaches highest (around 100\%) when the layer between convolutional and dense layer is chosen (i.e., flatten layer in VGG19 and GAP in MobileNetV1). 
This is because those layers contain both pixel and high-dimensional features, 
where neurons are more ideal for calculating neuron influence. 
When the last layer is chosen, DSR no longer increases. We speculate the reason is that nearly all information about classification is taken into consideration when flatten and GAP layer are chosen, so front and tail neurons are almost determined. With the chosen layer going deeper, DSR hardly changes.

\subsubsection{Influence of $\eta$ \label{eta}}
\begin{figure}[t]
\centering
    \subfigure[PGD]{
        \includegraphics[width=0.47\linewidth]{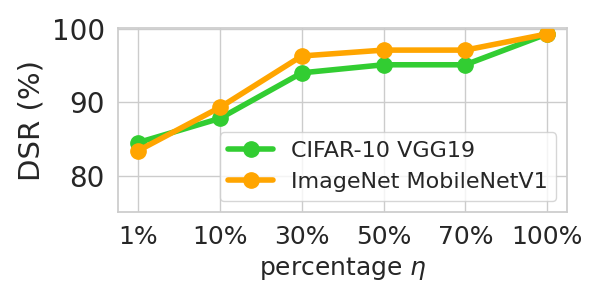}}
    \subfigure[PWA]{
        \includegraphics[width=0.47\linewidth]{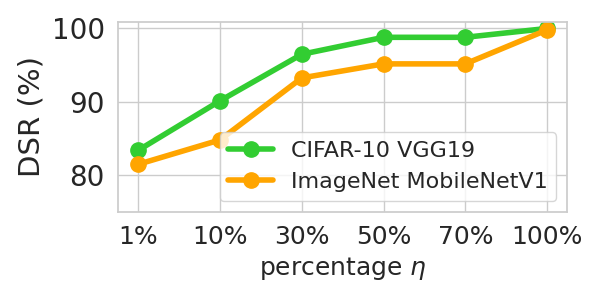}}
        \vspace{-10pt}
\caption{Results of DSR under different $\eta$.}
\label{candidate}
\vspace{-12pt}
\end{figure}

$\eta$ measures the percentage of the number of NIP candidates to that of unknown inputs. We calculate DSR on 1,000 adversarial examples from PGD and PWA under different $\eta$. Considering unsatisfactory situation, we set $\eta$ to 1\% in the worst case. This means that only 10 benign examples from training dataset are available for defenders. Results are shown in Fig. \ref{candidate}.

The number of NIP candidates shows little impact on defense results, e.g., even when $\eta$ is 1\%, DSR is around 82\%. This well indicates only a small number of inverse perturbations can cover various adversarial examples. 
Besides, DSR slightly increases with the growing of $\eta$. More suitable inverse perturbations can be found and added to adversarial examples by better-matched value of $\delta$, finally leading to better defense performance. Besides, comparatively higher DSR (around 95\%) and fewer NIP candidates can both be achieved when $\eta$=30\%, which meets the need of efficiency and effectiveness in the application.

As for baselines, defense methods related to feature mapping need more benign examples, e.g., 10 benign examples are far not enough for training Denoising Auto-Encoder in STL, or training EDSR in SR. Adversarial training strategies like CAS and SNS, also need numerous adversarial examples, which increases computation cost. Compared with baselines, NIP requires fewer benign examples but achieves higher DSR, so it is more light-weighted and less example-dependent. 

\subsection{Interpretation}
When interpreting effective defense, we refer to the following three aspects: $top$-$k$ neuron consistency (TKNC), visualizations of selected neurons and explanation of modifying inputs.

\subsubsection{Interpretation of Neuron Activation Consistency}
Theoretically, neurons with larger influence will be suppressed after attacks. After successful defense of reactive methods, neurons of defense examples will become similar to those of benign examples. We use $top$-$k$ neuron consistency (TKNC) to measure the consistency of important neurons, which is defined as follows:
\begin{equation} 
    TKNC(X,X',i,k)=\frac{top_k(X)\cap top_k(X') }{top_k(X)\cup top_k(X')}
\end{equation}
where $X$ denotes benign examples and $X'$ denotes corresponding defense examples. $top$ function selects $top$-$k$ important neurons in the $i$-th layer, calculated from $X$ and $X'$. Intuitively, a greater value of TKNC denotes higher consistency.

\begin{figure}[t]
\centering
\vspace{-5pt}
    \subfigure[VGG19, PGD, $\rho_s=0.714$]{
        \includegraphics[width=0.46\linewidth]{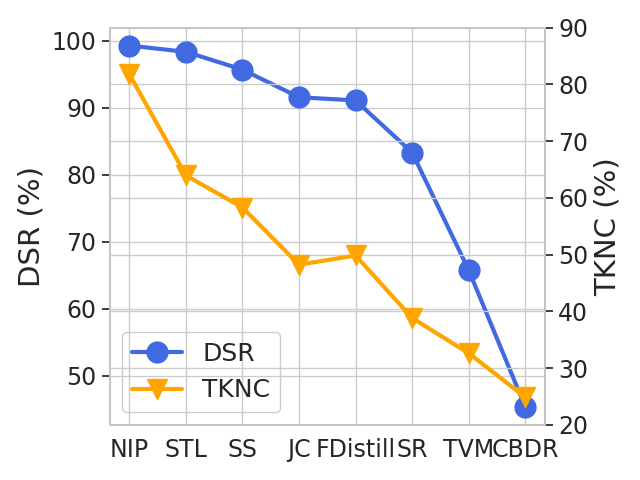}}
    \subfigure[VGG19, PWA, $\rho_s=1$]{
        \includegraphics[width=0.46\linewidth]{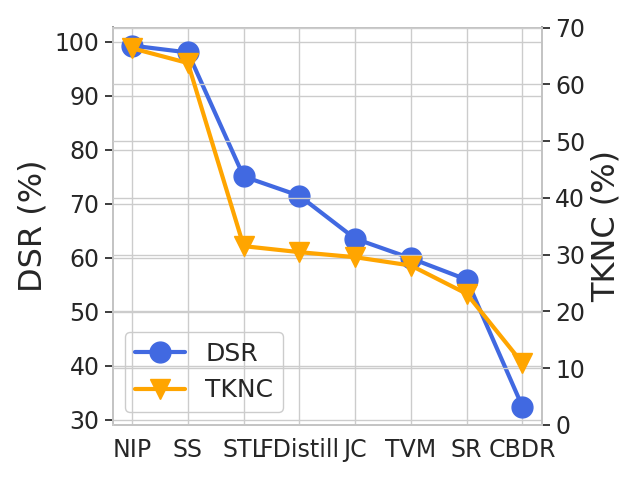}}\\
        \vspace{-10pt}
     \subfigure[MobileNetV1, PGD, $\rho_s=1$]{
        \includegraphics[width=0.46\linewidth]{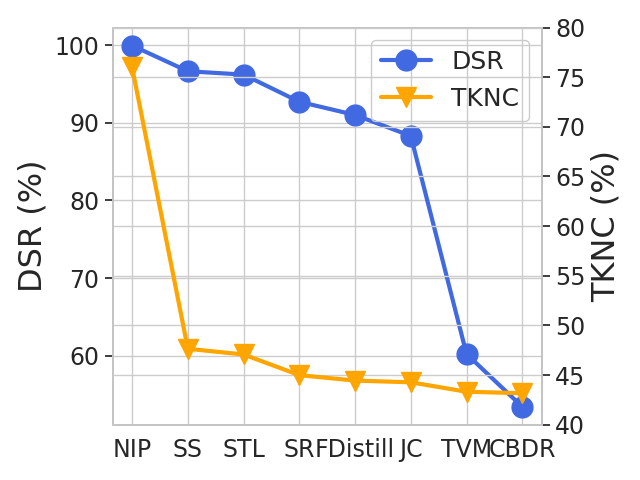}}
    \subfigure[MobileNetV1, PWA, $\rho_s=1$]{
        \includegraphics[width=0.46\linewidth]{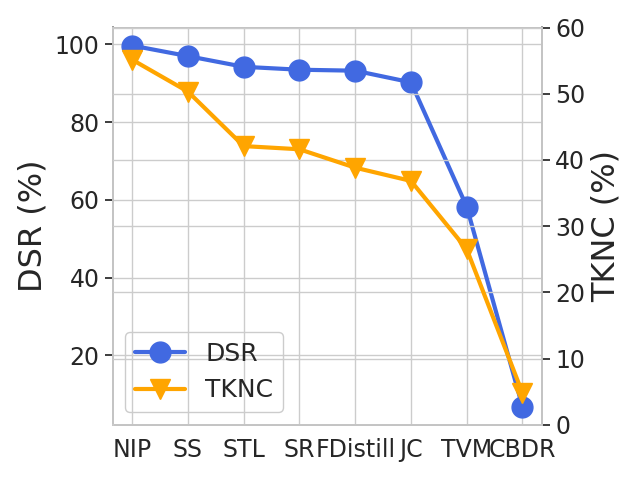}}
        \vspace{-10pt}
\caption{The relationship between DSR and TKNC. $\rho$ denotes the Spearman's rank correlation coefficient between DSR and TKNC.}
\label{TKNC}
\vspace{-8pt}
\end{figure}

\textbf{Implementation Details.} We further connect TKNC with DSR to demonstrate successful defenses and then make comparisons among four superior defense baselines on VGG19 of CIFAR-10 and MobileNetV1 of a-ImageNet. We calculate TKNC with $k$=5 on 1,000 adversarial examples crafted by PGD and PWA after defense. Results are shown in Fig. \ref{TKNC}. We also calculate the Spearman's rank correlation coefficient between DSR and TKNC, denoted as $\rho_s$ in the figure caption. Neuron indices used to calculate neuron influence are totally changed after retraining. So TKNC cannot be applied to interpret proactive defense (e.g., CAS, SNS and FDenoise). 

\textbf{Results and Analysis.} In most cases, defense performance of different reactive defense methods based on TKNC are consistent with those based on DSR. For instance, yellow and blue lines almost show the same trend: TKNC decreases when DSR drops. Besides, most $\rho$ values are 1, which indicates that TKNC and DSR is strongly correlated. Intuitively, defense examples with higher similarity of $top$-$k$ neuron distribution to that of benign ones, are more likely to be classified correctly. Therefore, we can leverage the $top$-$k$ neurons to determine and interpret successful defense performance.

It can be easily observed that TKNC of NIP is the highest (around 75\%) among all conducted defense baselines in all cases. This indicates that top neurons that most neurons that benign examples use for correct classification are utilized by NIP.
These neurons contribute most to correct classification. Thus, successful defense can be achieved when they are enhanced. After defense by NIP, models activate similar neurons as benign examples, so predictions will be correct. 

\begin{figure}[t]
\vspace{-5pt}
\centering
        \subfigure[VGG19]{
        \includegraphics[width=0.33\linewidth]{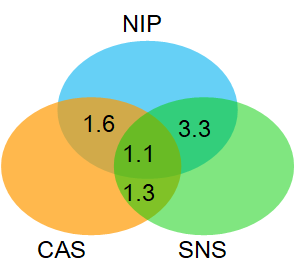}}
        \hspace{30pt}
    \subfigure[MobileNetV1]{
        \includegraphics[width=0.33\linewidth]{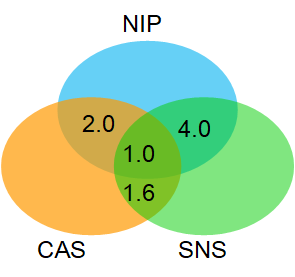}}
        \vspace{-10pt}
\caption{The overlap of selected neurons used in the defense.}
	\label{iou}
	\vspace{-12pt}
\end{figure}

\subsubsection{Interpretation of Selected Neurons}
We visualize selected neurons that used in the defense for better interpretation.

\textbf{Implementation Details.} We conduct experiments on VGG19 of CIFAR-10 and MobileNetV1 of a-ImageNet. We randomly select 100 benign examples for NIP and 100 PGD adversarial examples for CAS and SNS (for they are both PGD-based defenses), and calculate the overlap of selected neurons. For each defense, 10 neurons are selected in total. Specifically, $top$-10 neurons are chosen for CAS and SNS, while $top$-5 and $bottom$-5 are selected for NIP. Fig. \ref{iou} shows the results among NIP (blue), CAS (orange) and SNS (green). Numbers attached denoted the number of overlapped selected neurons on average.

\textbf{Results and Analysis.} It can be seen that three methods use similar neurons for defense. For instance, more than 60\% of $top$-10 neurons of NIP are overlapped with CAS and SNS, and these neurons are frequently chosen. But total neurons selected differ among defense. Based on general pattern of various attacks, NIP can target neurons that contribute to correct labels more straightforward. It deals with neurons in a more fine-grained level: front neurons are strengthened and tail ones are weakened. Thus it shows more effective defense against general adversarial perturbations.

\subsubsection{Interpretation of Modifying Inputs}
It may be quite counter-intuitive that we focus on the internal neurons, but the modification is done on input values. In this part, we try to explain the reason.

\begin{figure}[t]
\vspace{-7pt}
\centering
    \subfigure[$\text{NIP}_n$: activation values]{
        \includegraphics[width=0.455\linewidth]{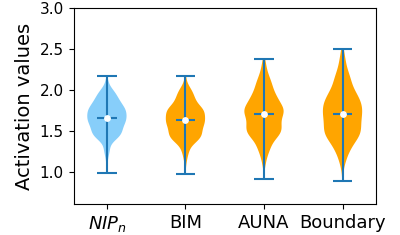}}
    \subfigure[NIP: pixel values]{
        \includegraphics[width=0.46\linewidth]{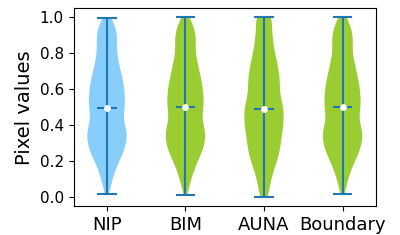}}
        \vspace{-10pt}
\caption{The distribution of candidates and different adversarial examples. Dataset: CIFAR-10; model: VGG19.}
\label{NIPn}
\vspace{-12pt}
\end{figure}

Recall that NIP adds inverse pixel perturbations on adversarial examples for defense. For comparison, we introduce $\text{NIP}_n$ that uses inverse perturbations on neuron activations in the chosen layer and modify the activation values of them in the test stage for defense. We further visualize neuron activation values of $\text{NIP}_n$ candidates and pixel values of NIP candidates in Fig. \ref{NIPn}. Average values of 100 adversarial examples per attack are represented and flatten layer is used to calculate neuron activation values. It is obvious that activation values are quite different among various attacks. On the contrary, input pixel values of different adversarial examples are all between [0,1] and their distributions are similar.  Therefore, compared with $\text{NIP}_n$, it's easier for NIP candidates to cover bounds of pixel values among adversarial examples, which finally leads to better defense performance. So we choose to modify inputs rather than directly modify neuron activation values. For more details and results, please refer to the \textbf{appendix}.

\section{Limitation and Discussion}
Although NIP has demonstrated its effectiveness of defending various adversarial attacks, it can be improved in the following aspects in the future. 

\textbf{Optimization of NIP Candidates.} As mentioned, the number of NIP candidates should be at least the same as the number of categories in the dataset. This shows limitations on the occasions that defenders should have the access to a number of benign training data before defense. Although we have done an experiment to demonstrate that NIP is likely still effective when only a few candidates are available, it can be seen that more candidates can guarantee better defense performance. We will work towards the trade-off between the number of candidates and defense effectiveness.

\textbf{Handling Attacks Not Based on Perturbations.} The effectiveness of NIP against universal adversarial attacks based on additive perturbation has been verified in the experiment. For attacks based on non-additive perturbation (e.g., spatial transformation), defense performance of NIP remains unsatisfied (with DSR around 50\%), which should be further improved in future work. NIP shows limitations on these attacks for its assumption that adversarial attacks are based on additive perturbations. It is possible to study more these attacks and feasibility of NIP against them.

\textbf{Better Efficiency.} Although NIP shows acceptable time cost in defense, it is still inferior to simple image transformations. The main reason goes to that the operation of neuron selection in a certain layer is time-consuming to some extent. Besides, unknown inputs are fed into the model twice (i.e., for neuron selection and reclassification), which increases the time cost. How to prune out unrelated neurons and how to directly target the most important neuron with larger influence deserve our efforts.

\section{Conclusions\label{Conclusions}}
In this paper, we introduce the concept of neuron influence and then divide neurons into front, tail and remaining neurons. We observe that multiple attacks fool the model by suppressing front neurons and enhancing tail ones. Motivated by it, we propose a new attack-agnostic defense method NIP, which in turn strengthens front neurons and weakening the tail part. As a result, it outperforms the state-of-the-art defense baselines on various datasets and models. Its effectiveness can also be verified on speaker recognition systems and online platforms. Besides, we also discuss adaptive attacks and provide some interpretable analysis for successful defense.










\bibliographystyle{IEEEtran}
\bibliography{ref}

\begin{IEEEbiography}[{\includegraphics[width=1in,height=1.2in,clip,keepaspectratio]{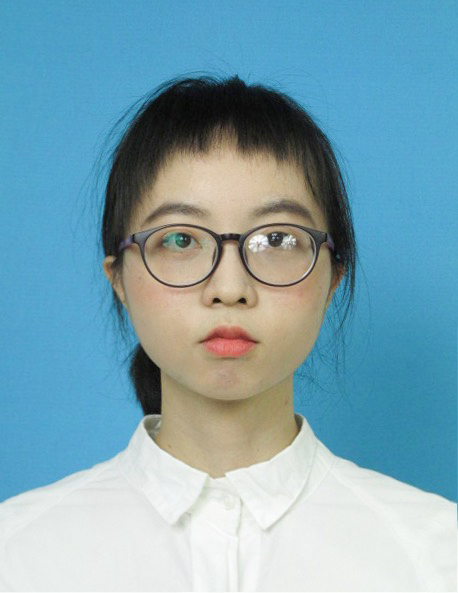}}]{Ruoxi Chen} is currently a PhD student with the Institute of Information Engineering, Zhejiang University of Technology, Hangzhou, China. Her research interest covers trustworthy machine learing and computer vision.
\end{IEEEbiography}

\begin{IEEEbiography}[{\includegraphics[width=1in,height=1.2in,clip,keepaspectratio]{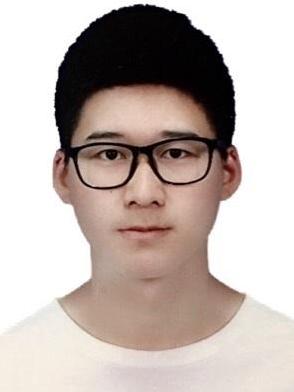}}]{Haibo Jin} is a PhD student in the Institute of Information Engineering, Zhejiang University of Technology, Hangzhou, China. His research interest include artificial intelligence, and adversarial attack and defense.
\end{IEEEbiography}

\begin{IEEEbiography}[{\includegraphics[width=1in,height=1.2in,clip,keepaspectratio]{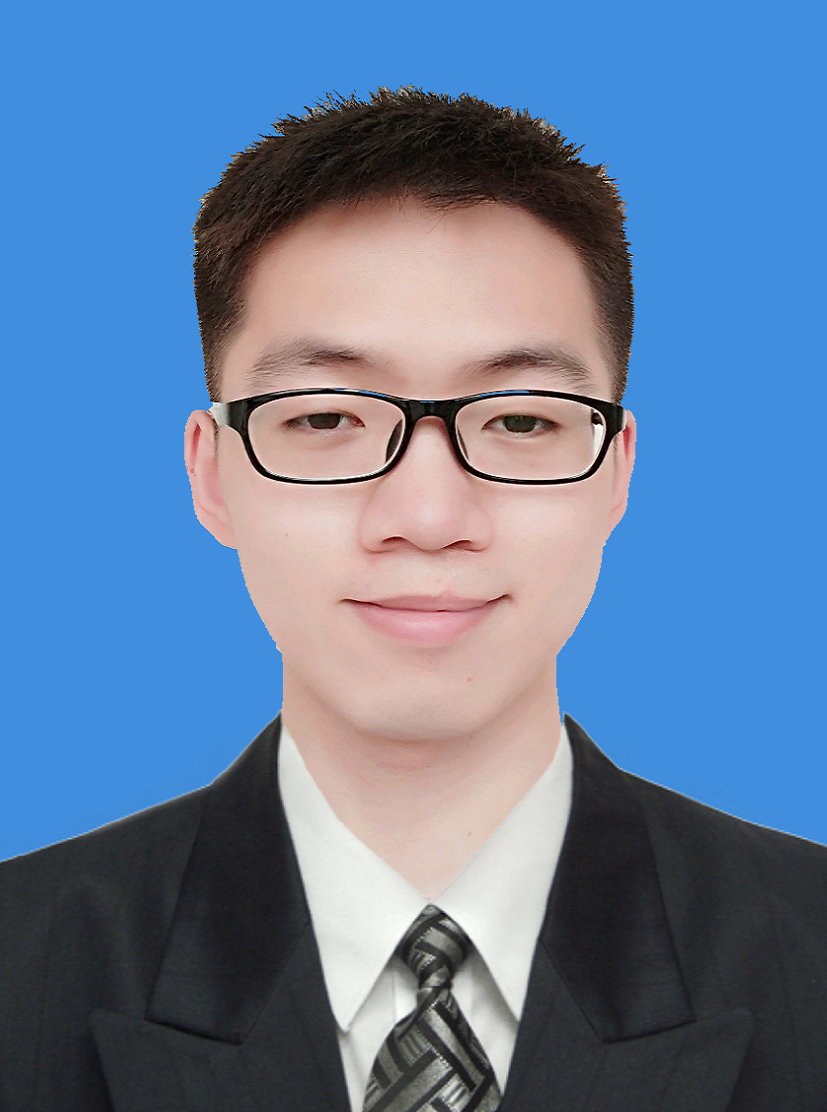}}]{Haibin Zheng} received B.S. and Ph.D. degrees from Zhejiang University of Technology, Hangzhou, China, in 2017 and 2022, respectively. He is currently a university lecturer at the Institute of Cyberspace Security, Zhejiang University of Technology. His research interests include deep learning and artificial intelligence security.
\end{IEEEbiography}

\begin{IEEEbiography}[{\includegraphics[width=1in,height=1.2in,clip,keepaspectratio]{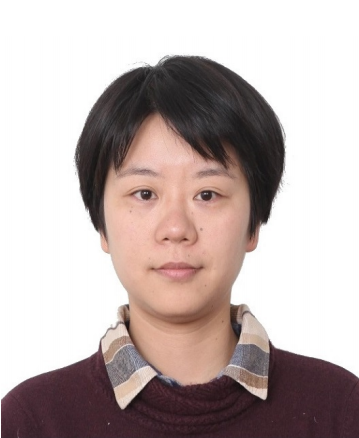}}]{Jinyin Chen} received BS and PhD degrees from Zhejiang University of Technology, Hangzhou, China, in 2004 and 2009, respectively. She studied evolutionary computing in Ashikaga Institute of Technology, Japan in 2005 and 2006. She is currently a professor in the Institute of Cyberspace Security and the College of Information Engineering, Zhejiang University of Technology. Her research interests include artificial intelligence security, graph data mining and evolutionary computing.
\end{IEEEbiography}

\begin{IEEEbiography}[{\includegraphics[width=1in,height=1.2in,clip,keepaspectratio]{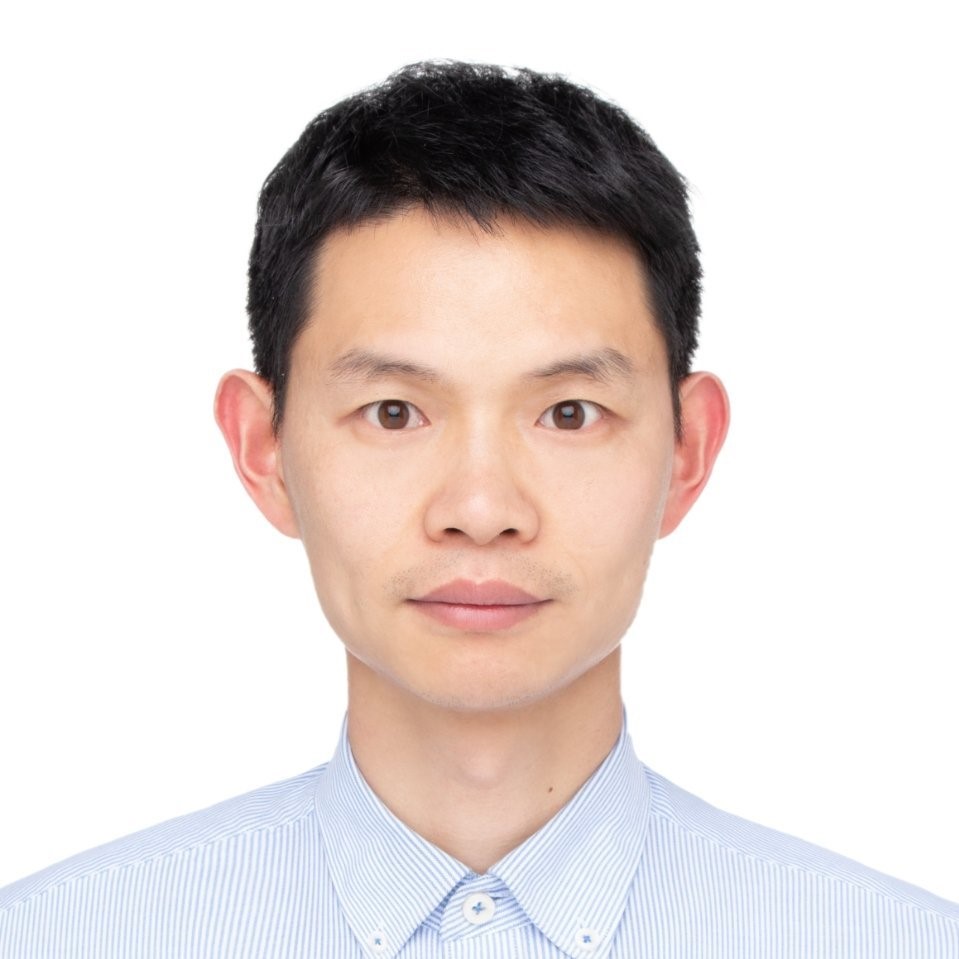}}]{Zhenguang Liu} is currently a research fellow in Zhejiang University. He had been a research fellow in National University of Singapore and A*STAR (Agency for Science, Technology and Research, Singapore) for several years. He respectively received his Ph.D. and B.E. degrees from Zhejiang University and Shandong University, China. His research interests include multimedia and security.
\end{IEEEbiography}

\end{document}


\appendices

\section{Details of Experiment Setup \label{appendix_setup_details}}

ASR of adversarial examples are shown in Tables~\ref{ASR}.  
The attack parameter settings are shown in Table~\ref{tab:Parameter_Setting_attack}. ``epsilon'' is the maximum perturbation size of the given example. ``stepsize'' is the perturbation step size of one iteration.
The implementations of FGSM~\cite{GoodfellowSS14}, BIM~\cite{Kurakin2017Adversarial}, MI-FGSM~\cite{dong2018boosting}, JSMA~\cite{papernot2016limitations}, PGD~\cite{Madry2018Towards}, DeepFool~\cite{moosavi2016deepfool}, AUNA~\cite{FoolboxAUNA}, PWA~\cite{SchottRBB19}, Boundary~\cite{Brendel2018Boundary} are from the foolbox tool\footnote{The code of attacks can be downloaded at: https://
foolbox.readthedocs.io/en/v2.3.0/modules/attacks.html}. UAP~\cite{moosavi2017universal} and AutoAttack~\cite{croce2020reliable} are conducted on adversarial robustness toolbox\footnote{https://github.com/Trusted-AI/adversarial-robustness-toolbox/tree/main/art}. We obtained the implementation of EOT~\cite{athalye2018synthesizing} and One pixel (Pixel)~\cite{su2019one} from GitHub\footnote{https://github.com/prabhant/synthesizing-robust-adversarial-examples}\footnote{https://github.com/Hyperparticle/one-pixel-attack-keras}, respectively.

\begin{table}[htbp]
\Huge
\centering
\caption{ASR of adversarial examples.}
\resizebox{1\linewidth}{!}{
\begin{tabular}{cccccccccccc}
\toprule
\multicolumn{2}{c}{\textbf{Datasets}} &
  \multicolumn{2}{c}{CIFAR-10} &
  \multicolumn{2}{c}{CIFAR-100} &
  \multicolumn{2}{c}{GTSRB} &
  \multicolumn{2}{c}{ImageNet} &
  \multicolumn{2}{c}{VCTK} \\ \cline{3-12} 
\multicolumn{2}{c}{\textbf{Models}} &
  VGG19 &
  AlexNet &
  DenseNet &
  SqueezeNet &
  LeNet-5 &
  ResNet20 &
  VGG19 &
  MobileNetV1 &
  ResNet34 &
  DeepSpeaker \\ \hline
\multirow{13}{*}{\textbf{\rotatebox{90}{Attacks}}} &
  FGSM &
  100.00\% &
  99.52\% &
  100.00\% &
  100.00\% &
  99.62\% &
  100.00\% &
  99.97\% &
  100.00\% &
  97.70\% &
  99.42\% \\
 & BIM      & 99.87\%  & 96.18\% & 100.00\% & 100.00\% & 97.58\%  & 99.95\%  & 99.93\% & 99.95\%  & 97.70\% & 99.42\%  \\
 & MIFGSM   & 99.76\%  & 97.48\% & 100.00\% & 100.00\% & 100.00\% & 99.84\%  & 99.67\% & 99.97\%  & 97.70\% & 99.42\%  \\
 & JSMA     & 100.00\% & 99.88\% & 100.00\% & 100.00\% & 99.66\%  & 99.68\%  & 99.87\% & 99.97\%  & 97.70\% & 99.42\%  \\
 & PGD      & 99.92\%  & 95.98\% & 100.00\% & 100.00\% & 99.95\%  & 99.91\%  & 99.90\% & 99.97\%  & 97.70\% & 99.42\%  \\
 & DeepFool & 99.66\%  & 96.64\% & 100.00\% & 100.00\% & 94.56\%  & 99.63\%  & 99.35\% & 99.82\%  & 97.70\% & 99.42\%  \\
 & UAP      & 86.02\%  & 75.80\% & 87.20\%       & 88.60\%  & 82.40\% & 88.30\% & 91.15\% & 94.15\%  & 97.70\% & 99.42\%  \\
 & EOT      & 88.10\%  & 85.80\% & 93.10\%       & 89.90\%  & 88.90\% & 89.86\% & 87.25\% & 86.00\%  & / & /  \\
 & AUNA     & 100.00\% & 99.98\% & 100.00\% & 100.00\% & 99.95\%  & 99.95\%  & 99.95\% & 100.00\% & 97.70\% & 99.42\%  \\
 & PWA      & 97.60\%  & 87.32\% & 98.00\%  & 98.50\%  & 99.85\%  & 99.65\%  & 87.75\% & 90.00\%  & 98.25\% & 100.00\% \\
 & Pixel    & 100.00\% & 99.97\% & /        & /        & /        & /        & /       & /        & /       & /        \\
 & Boundary & 93.00\%  & 95.60\% & 89.50\%  & 88.90\%  & 85.80\%  & 83.40\%  & 87.50\% & 89.25\%  & 83.25\% & 83.25\%  \\
  & Auto     & 84.50\%  & 78.80\% & 88.20\%  & 90.00\%  & 85.95\% & 86.74\% & 89.25\% & 92.75\%  & / & /  \\
 \bottomrule
\end{tabular}}
\label{ASR}
\vspace{-10pt}
\end{table}

\begin{table}[htbp]
\Huge
    \centering
    \caption{The attack parameter settings of adversarial attacks.
    }
    \resizebox{\linewidth}{!}{
        \begin{tabular}{ll}
        \toprule 
        \multicolumn{1}{l}{\textbf{Attacks}} & \multicolumn{1}{l}{\textbf{Parameter Setting}}                                     \\ \hline
        FGSM                                 & $l_\infty$ epsilon=0.2, max\_iteration=1000.                                                      \\
        BIM                         & $l_\infty$ epsilon=0.3, stepsize=0.05, iterations=10. \\
        MI-FGSM                              & $l_\infty$ epsilon=0.3, iterations=10, stepsize=0.06, decay\_factor=0.9.                      \\
        JSMA                               & $l_0$ epsilon=0.1, max\_iter=2000,max\_perturbations\_per\_pixel=7.                                                     \\
        PGD                                  & $l_\infty$ epsilon=0.3, stepsize=0.01,  iterations=40.     \\
        DeepFool                                 & $l_2$ epsilon=0.3, steps=100, subsample=10.                                                    \\
        UAP                                 & $l_\infty$ epsilon=0.1, max\_iter=3.                                                      \\
        EOT                            &  $l_2$ epsilon=0.05, iterations=5.  \\
        AUNA                           & $l_2$ epsilon=0.2, max\_iter=1000.                                                     \\
        PWA                                 & $l_2$ epsilon=0.3, max\_iter=1000.                                         \\
        Pixel                               &   modify the image by making one pixel yellow.   \\
        Boundary                             & $l_2$ epsilon=0.2, iterations=5,000, max\_directions=25, initial\_stepsize=1e-2.\\
        Auto                           &  FGSM($l_\infty$ epsilon=0.2), PGD($l_\infty$ epsilon=0.2, stepsize=0.1), UAP($l_\infty$ epsilon=0.1, max\_iter=3). \\
         \bottomrule
        \end{tabular}
            \label{tab:Parameter_Setting_attack}
            \vspace{-5pt}
    } 
\end{table}

The defense settings of image classification and speaker recognition are shown in Table~\ref{defense image baseline} and Table \ref{defense audio baseline}, respectively. 

The implementations of SS~\cite{Xu0Q18}, CBDR~\cite{Xu0Q18}, JC~\cite{dziugaite2016study} and TVM~\cite{guo2018countering} are from adversarial robustness toolbox. We obtained the source code of STL~\cite{sun2019adversarial}, FDistill~\cite{liu2019feature}, FDenoise~\cite{xie2019feature}, SR~\cite{mustafa2019image} and CAS~\cite{bai2020improving} from GitHub\footnote{https://github.com/GitBoSun/AdvDefense\_CSC}\footnote{https://github.com/sibosutd/feature-distillation}\footnote{https://github.com/facebookresearch/
ImageNet-Adversarial-Training}\footnote{https://
github.com/aamir-mustafa/super-resolution-adversarial-defense}\footnote{https://github.com/bymavis/CAS\_ICLR2021}. They are configured according to the best performance setting reported in the respective papers. We implement our copy of SNS~\cite{zhang2020interpreting} following the introduced technical approach. 

We implement Down sampling (DS)~\cite{YangLCS19}, Random crop (RC)~\cite{li2006localized}, MP3 compression (MC)~\cite{andronic2020mp3} and Quantization (QTZ)~\cite{YangLCS19} for baselines of speaker recognition.

\begin{table}[htbp]
\centering
\caption{The default settings of defense baselines in image classification.}
\resizebox{0.9\linewidth}{!}{
\begin{tabular}{ll}
\toprule 
\multicolumn{1}{l}{\textbf{Defense}} & \multicolumn{1}{l}{\textbf{Parameter Settings}}                                     \\ \hline
        SS                  & window\_size=2 \\          
        JC                       & quality=80                     \\   
        TVM                & prob=0.5, $\lambda$=0.1     \\
        STL              & K=64, P=8, lmbda = 0.1, npd=16, fltlmbd=5  \\
        FDistill           & qs=40, factor=100, block\_side=8         \\
        FDenoise           & median filter                  \\
        SR                  & EDSR network(scale=4, num\_res\_blocks=16), sigma=0.02 \\ \bottomrule
        \end{tabular}
    }
    \label{defense image baseline}
    \vspace{-10pt}
\end{table}

\begin{table}[htbp]
\centering
\caption{The default settings of defense baselines in speaker recognition.}
\resizebox{1\linewidth}{!}{
\Large
\begin{tabular}{ll}
\toprule
\textbf{Defense} & \textbf{Parameter Settings}                                          \\ \hline
DS &
  \begin{tabular}[c]{@{}l@{}}down-samples a band-limited audio file without sacrificing the\\ quality of the recovered signal. We set $down\_sample\ rate$=8kHz.\end{tabular} \\ \cline{2-2} 
RC               & randomly crops data points from the signal. We set $crop num$=1,000. \\ \cline{2-2} 
MC &
  \begin{tabular}[c]{@{}l@{}}conducts compress operations to audio adversarial examples \\ to decrease the impact of adversarial noise. We set $sample\ rate$=16k.\end{tabular} \\ \cline{2-2} 
QTZ &
  \begin{tabular}[c]{@{}l@{}}rounds the amplitude of audio sampled data into the nearest \\ integer multiple, for reducing the adversarial perturbations. We set $q$=256.\end{tabular} \\\bottomrule
\end{tabular}}
    \label{defense audio baseline}
    \vspace{-10pt}
\end{table}

\section{More Results of Adaptive Attacks}
\subsection{Other Types of Adaptive Attacks}
We also design three types of adaptive attacks: secondary adversarial attack (SAA), adaptive loss attack (ALA) and iterative adaptive loss attack (I-ALA).

(1) SAA. Attackers use the same attack approach to add adversarial perturbations again on examples after successful defense by NIP. Then these adversarial examples after second-time attack will be fed into the model for classification and ASR will be calculated. We design this type of attack to further demonstrate the robustness of inverse perturbations by NIP. 

(2) ALA. ALA directly strengthens tail neurons and weakens front ones in turn. Specifically, adversarial examples are generated as follows:
\begin{equation}
\begin{aligned}
    &loss_x= \sum_{j=0}^{k_1}\varphi_i(x)-\sum_{i=0}^{k_2}\varphi_j(x) \\
    &x^*= x+ s \cdot \nabla_x loss_x
\end{aligned}
\label{ala}
\end{equation}
where $x^*$ denotes the generated adversarial example when the benign example $x$ is input. $k_1$ is responsible for front neurons and $k_2$ for tail ones. $s$ controls the perturbation size, which is set to 0.01. $\varphi_i(x)$ denotes the output of the $i$-th neuron in the front neuron $\Omega_f$ when fed with input $x$.

(3) I-ALA. On the basis of ALA, we use a multi-step attack to enhance its effectiveness. We formulate the attacks as follows: 
\begin{equation}
    x^*_{m+1}=x^*_{m}+ s\cdot \nabla_x loss_x
\end{equation}
$loss_x$ is calculated according to Eq. \ref{ala}. Here, $s$ is set to 0.01 and the iteration number $m$ is 20.

\textbf{Implementation Details.} 
(1) Experiments are conducted on three models, AlexNet~\cite{krizhevsky2012imagenet} of CIFAR-10~\cite{krizhevsky2009learning}, ResNet20~\cite{he2016deep} of GTSRB~\cite{Stallkamp-IJCNN-2011} and VGG19~\cite{simonyan2014very} of a-ImageNet.  
(2) For SAA, FGSM, PGD, PWA, Boundary are adopted. For ALA and I-ALA, $k_1$=$k_2$=3.
(3) For defense, $k_1$=$k_2$=3 for ALA and I-ALA. As for defending others, $k_1$=$k_2$=5. For measurement, ASR of these adaptive attacks are calculated on 1,000 adversarial examples. Here, DSR that measures defense performance is defined as: $DSR=1-ASR$.  
(4) Results of ASR are shown in Table \ref{adaptive_asr}, where FGSM$^*$ means SAA by FGSM against the given defense by NIP, etc.

\begin{table}[t]
\huge
\centering
\caption{ASR of adaptive attacks.}
\resizebox{0.9\linewidth}{!}{
\begin{tabular}{cccccccc}
\toprule
\multirow{2}{*}{\textbf{Datasets}} & \multirow{2}{*}{\textbf{Models}} & \multicolumn{6}{c}{\textbf{Attacks}} \\ \cline{3-8} 
         &          & FGSM$^*$  & PGD$^*$   & PWA$^*$   & Boundary$^*$ & ALA     & I-ALA      \\ \hline
CIFAR-10 & AlexNet  & 3.55\% & 1.91\% & 4.58\% & 0.28\%    & 10.30\% & 14.70\%  \\
GTSRB    & ResNet20 & 0.31\% & 0.00\% & 0.76\% & 0.60\%    & 18.50\% & 6.40\%   \\
a-ImageNet & VGG19    & 2.32\% & 3.26\% & 1.22\% & 0.91\%    & 20.40\% & 12.70\%  \\ \bottomrule
\label{adaptive_asr}
\end{tabular}}
\vspace{-15pt}
\end{table}

\textbf{Results and Analysis.} From Table \ref{adaptive_asr}, we can observe that most of these adaptive attacks are invalid against NIP. This indicates that when faced with adaptive attacks, NIP can still achieve high defense performance.

For SAA, NIP can alleviate adversarial effect by accurately identifying tail neurons that will be utilized by attacks. Thus, the inverse perturbation added on the example can offset the adversarial effect in advance, which makes it harder to successfully conduct second-time attacks.

ASR of ALA is lower than 21\%, which shows that NIP can still successfully defend a majority of adversarial examples on three models. When ALA becomes a multi-step attack I-ALA, the adversarial effect doesn't show significant increase. The defense of NIP remains stable. For instance, DSR of ALA on AlexNet is 89.70\% while that of I-ALA is 85.30\%. This indicates the adversarial effect can be offset by NIP, based on front neurons that related to correct predictions and tail ones that are utilized by general attacks.

\subsection{More Results of S-ALA}
In this section, we report the results of S-ALA on different $\lambda$ and iterations. 
Experiments are conducted on three models, AlexNet of CIFAR-10, ResNet20 of GTSRB and VGG19 of ImageNet. We generate 1000 adversarial examples of S-ALA and measure the effectiveness by ASR. $k$ of NIP is set to 5. Neuron influence of each neuron on flatten layer or global average pooling is used to calculate $\sigma$ for each example. 

\begin{figure}[htbp]
\renewcommand{\@thesubfigure}{\hskip\subfiglabelskip}
\makeatother
\subfigbottomskip=-3pt
\centering
    \subfigure[(a) $\lambda$=1, iter=10]{
        \includegraphics[width=0.46\linewidth]{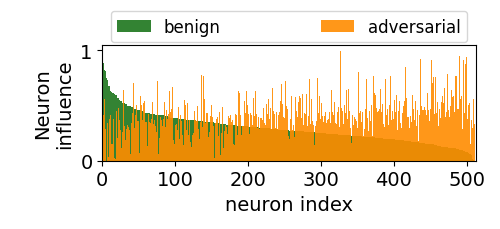}}
    \subfigure[(d) $\lambda$=0.8, iter=10]{
        \includegraphics[width=0.46\linewidth]{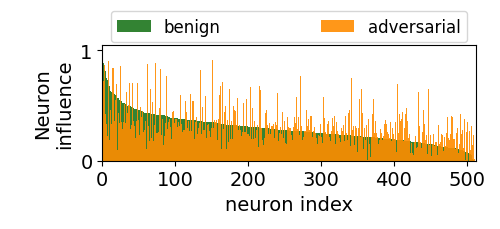}}\\
    \subfigure[(b) $\lambda$=1, iter=15]{
        \includegraphics[width=0.46\linewidth]{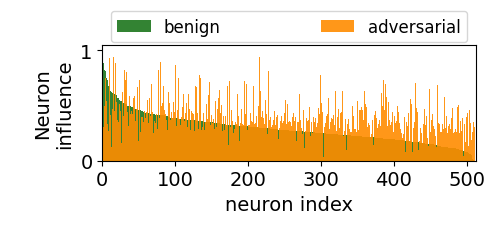}}
    \subfigure[(e) $\lambda$=1, iter=10]{
        \includegraphics[width=0.46\linewidth]{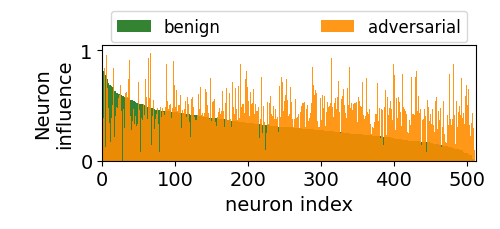}}\\ 
    \subfigure[(c) $\lambda$=1, iter=30]{
        \includegraphics[width=0.46\linewidth]{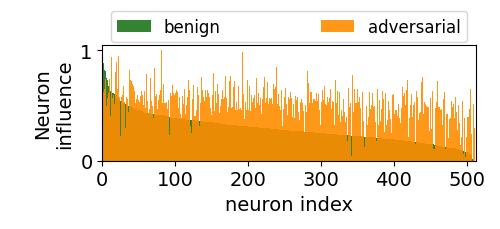}}
    \subfigure[(f) $\lambda$=1.2, iter=10]{
        \includegraphics[width=0.46\linewidth]{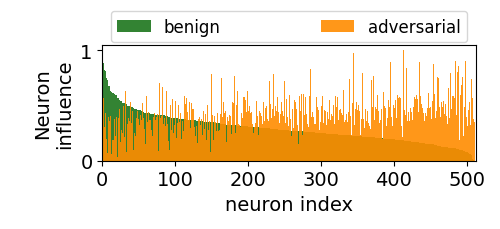}}\\
\caption{Neuron influence on benign (green) and adversarial examples of S-ALA (orange). ``iter" denotes iteration numbers. }
\label{S-ALA}
\vspace{-10pt}
\end{figure}

Besides, we also visualize the change of neuron influence on S-ALA, with respect to different iterations and $\lambda$. 500 randomly-chosen adversarial examples are used. Flatten layer of VGG19 on CIFAR-10 with 512 neurons are used for illustration. Results are shown in Fig \ref{S-ALA}.

\begin{figure}[htbp]
\centering
    \subfigure[Influence of iterations]{
        \includegraphics[width=0.46\linewidth]{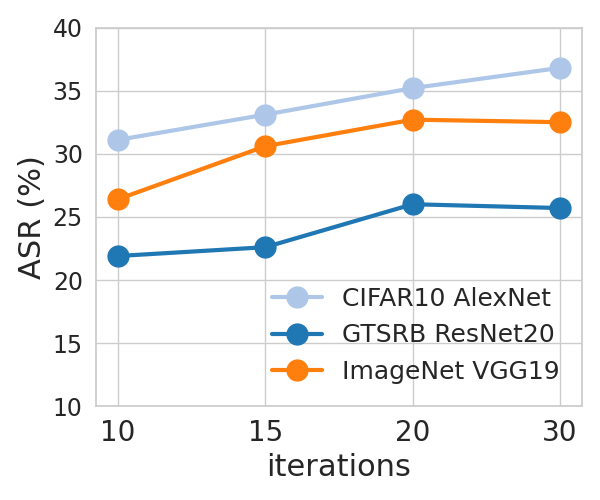}}
    \subfigure[Influence of $\lambda$]{
        \includegraphics[width=0.46\linewidth]{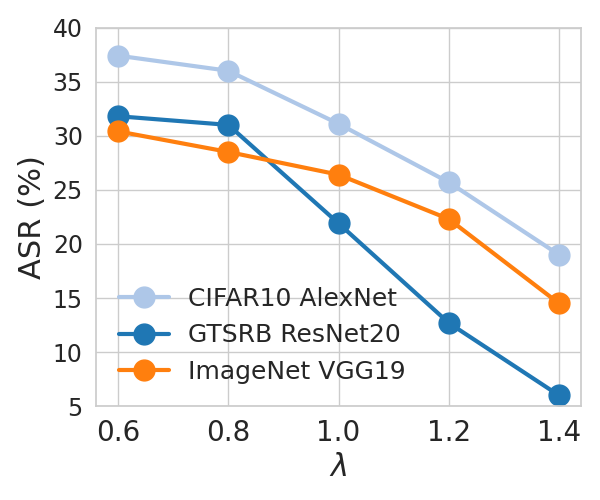}}
 \vspace{-5pt}       
\caption{Parameter sensitivity of S-ALA on different iterations and $\lambda$.}
\label{para_S-ALA}
\vspace{-10pt}
\end{figure}
 
\makeatletter

\subsubsection{Influence of Iterations}
We iterated the calculation of each adversarial example for 10, 15, 20 and 30 rounds, to further evaluate the effectiveness of NIP on S-ALA. $\lambda$ is set to 1 by default. Results are shown in Fig. \ref{para_S-ALA}(a).

ASR of S-ALA gradually increases as iterations grow. This is because with respect to neuron influence, adversarial examples generated by S-ALA are more similar to benign examples. So it is harder for NIP to target front and tail neurons to offset adversarial effect. When iterations are set to 30, ASR of it remains stable. We speculate the possible reason that almost all front and tail neurons are manipulated the attack so ASR of it no longer changes.

It can be observed from (a), (b), (c) in Fig. \ref{S-ALA} that as iterations increase, the neuron influence of S-ALA first approximates and then exceeds that of benign examples. In this process, ASR of S-ALA gradually increases. But front and tail neurons found by NIP are still responsible for correct and erroneous classification. By in turn strengthening front neurons and suppressing tail ones, defense can still be achieved.

\subsubsection{Influence of $\lambda$}
$\lambda$ is the balancing parameter between similarity of neuron influence and misclassification. Here, we generate S-ALA examples for 10 rounds of iterations. Results are shown in Fig. \ref{para_S-ALA}(b).

ASR of S-ALA drops with the increase of $\lambda$. Theoretically, larger $\lambda$ means smaller similarity in neuron influence between benign and adversarial examples. So neuron influence of adversarial examples generated with larger $\lambda$ is more similar with that of general adversarial examples. This makes it easier for NIP to offset the adversarial effect. Thus, ASR of S-ALA decreases.

In (d), (e), (f) in Fig. \ref{S-ALA}, as $\lambda$ increases, neuron influence of adversarial examples generated by S-ALA shows upward trend, which approximates that of other general adversarial attacks. Front and tail neurons are easier to target for NIP so more than 70\% adversarial examples of S-ALA fail to fool the model. 

\section{More Visualizations}

\subsection{Visualizations of Neuron Changes\label{neuron changes}}
\begin{figure}[htbp]
\renewcommand{\@thesubfigure}{\hskip\subfiglabelskip}
\makeatother
\subfigbottomskip=-3pt
\centering
    \subfigure[]{
        \includegraphics[width=0.46\linewidth]{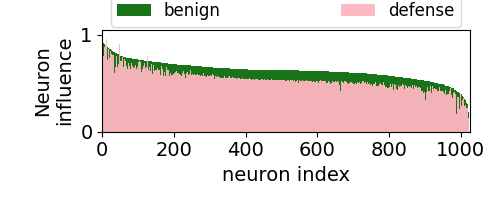}}
    \subfigure[]{
        \includegraphics[width=0.46\linewidth]{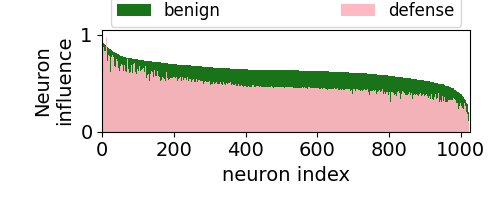}}\\
    \subfigure[]{
        \includegraphics[width=0.46\linewidth]{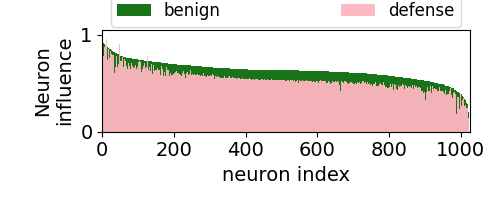}}
    \subfigure[]{
        \includegraphics[width=0.46\linewidth]{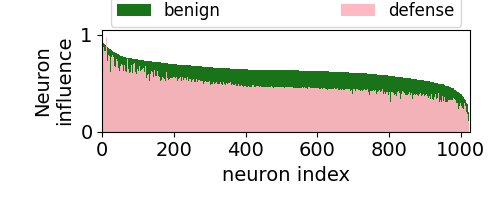}}\\ 
    \subfigure[(a) STL]{
        \includegraphics[width=0.46\linewidth]{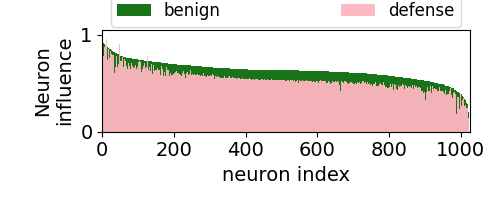}}
    \subfigure[(b) NIP]{
        \includegraphics[width=0.46\linewidth]{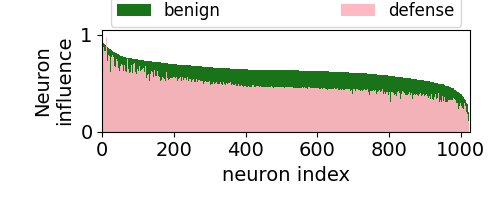}}\\
\caption{Changes of neuron influence on benign (green) and adversarial examples (pink). Adversarial examples are generated by DeepFool (Row 1), FGSM (Row 2) and Boundary (Row 3).}
\label{neuron-defense}
\vspace{-8pt}
\end{figure}

We visualize changes of neuron influence on benign and defense examples after defense in Fig. \ref{neuron-defense}. The left column shows defense after STL while the right represents that after NIP. 500 adversarial examples per attack and global\_average\_pooling of MobileNetV1 on a-ImageNet, with 1024 neurons, are used for illustration. The left column shows defense after STL while the right represents that after NIP.

\makeatletter

\makeatletter
\begin{figure}[t]
\renewcommand{\@thesubfigure}{\hskip\subfiglabelskip}
\makeatother
\subfigbottomskip=-13pt
\setlength{\abovecaptionskip}{16pt}

\centering
    \subfigure[]{
        \includegraphics[width=0.15\linewidth]{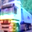}}
    \subfigure[]{
        \includegraphics[width=0.15\linewidth]{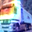}}
    \subfigure[]{
        \includegraphics[width=0.15\linewidth]{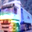}}
    \subfigure[]{
        \includegraphics[width=0.15\linewidth]{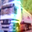}}
    \subfigure[]{
        \includegraphics[width=0.15\linewidth]{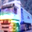}}\\
 \hspace{-1.5pt}       
    \subfigure[]{
        \includegraphics[width=0.15\linewidth]{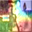}}
    \subfigure[]{
        \includegraphics[width=0.15\linewidth]{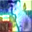}}
    \subfigure[]{
        \includegraphics[width=0.15\linewidth]{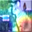}}
    \subfigure[]{
        \includegraphics[width=0.15\linewidth]{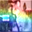}}
    \subfigure[]{
        \includegraphics[width=0.15\linewidth]{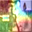}}\\
    \subfigure[benign]{
        \includegraphics[width=0.15\linewidth]{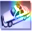}}
    \subfigure[PGD]{
        \includegraphics[width=0.15\linewidth]{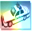}}
    \subfigure[PDG-def]{
        \includegraphics[width=0.15\linewidth]{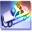}}
    \subfigure[BD]{
        \includegraphics[width=0.15\linewidth]{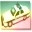}}
    \subfigure[BD-def]{
        \includegraphics[width=0.15\linewidth]{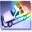}}\\
 \hspace{-1.5pt}   
    \subfigure[]{
        \includegraphics[width=0.15\linewidth]{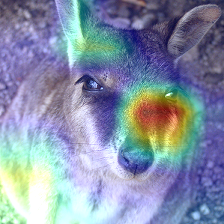}}
    \subfigure[]{
        \includegraphics[width=0.15\linewidth]{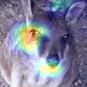}}
    \subfigure[]{
        \includegraphics[width=0.15\linewidth]{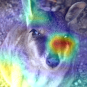}}
    \subfigure[]{
        \includegraphics[width=0.15\linewidth]{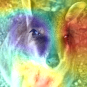}}
    \subfigure[]{
        \includegraphics[width=0.15\linewidth]{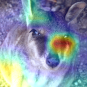}} \\
    \subfigure[]{
        \includegraphics[width=0.15\linewidth]{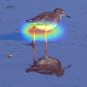}}
    \subfigure[]{
        \includegraphics[width=0.15\linewidth]{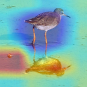}}
    \subfigure[]{
        \includegraphics[width=0.15\linewidth]{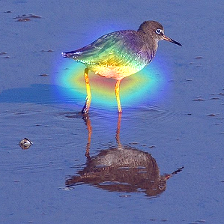}}
    \subfigure[]{
        \includegraphics[width=0.15\linewidth]{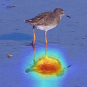}}
    \subfigure[]{
        \includegraphics[width=0.15\linewidth]{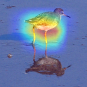}}\\
    \hspace{0.5pt}
    \subfigure[benign]{
        \includegraphics[width=0.15\linewidth]{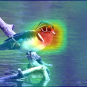}}
    \subfigure[PGD]{
        \includegraphics[width=0.15\linewidth]{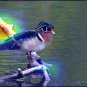}}
    \subfigure[PDG-def]{
        \includegraphics[width=0.15\linewidth]{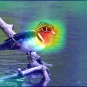}}
    \subfigure[BD]{
        \includegraphics[width=0.15\linewidth]{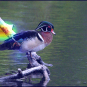}}
    \subfigure[BD-def]{
        \includegraphics[width=0.15\linewidth]{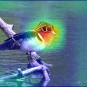}}
\caption{Grad-CAM visualizations of benign examples, adversarial examples
crafted by PGD and Boundary and defense examples after NIP, where Boundary is abbreviated as ``BD". }
\label{gradcam}
\end{figure}
\subsection{Visualizations Based on Grad-CAM}
We focus on attention map to visualize the feature learned by models, to interpret successful defense.

\textbf{Implementation Details.} (1) We use Grad-CAM~\cite{selvaraju2017grad} to visualize the attention map changes between benign, adversarial examples and defense examples after NIP. Areas that are more class-relevant and drawn more attention by the model are painted in red. The weight that the model allocates decreases from red to blue. (2) The visualization of VGG19 on CIFAR-10 and MobileNetV1 on a-ImageNet are shown in Fig. \ref{gradcam}. First three rows represent examples from CIFAR-10 while last three rows show those from a-ImageNet.

\textbf{Results and Analysis.} In the first column, red areas in benign heatmaps mainly located in the object related to class label, thus images are correctly classified. On the contrary, when model focuses on areas irrelevant to the label, misclassification occurs, as shown in heatmaps of PGD and Boundary in the second and fifth columns. Similarities of red areas can be observed in defense examples and benign ones. This demonstrates that NIP could rectify classification results by strengthening class-relevant neurons while weakening irrelevant ones. The effect of adversarial perturbations is nearly eliminated so right labels are obtained after defense.

\makeatletter
\begin{figure}[t]
\subfigbottomskip=-13pt
\renewcommand{\@thesubfigure}{\hskip\subfiglabelskip}
\makeatother
\setlength{\abovecaptionskip}{16pt}
\centering
    \subfigure[]{
        \includegraphics[width=0.17\linewidth]{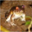}}
    \subfigure[]{
        \includegraphics[width=0.17\linewidth]{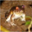}}
    \subfigure[]{
        \includegraphics[width=0.17\linewidth]{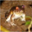}}
    \subfigure[]{
        \includegraphics[width=0.17\linewidth]{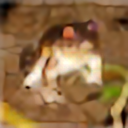}}
    \subfigure[]{
        \includegraphics[width=0.17\linewidth]{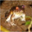}}\\
    \subfigure[]{
        \includegraphics[width=0.17\linewidth]{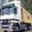}}
    \subfigure[]{
        \includegraphics[width=0.17\linewidth]{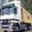}}
    \subfigure[]{
        \includegraphics[width=0.17\linewidth]{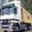}}
    \subfigure[]{
        \includegraphics[width=0.17\linewidth]{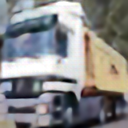}}
    \subfigure[]{
        \includegraphics[width=0.17\linewidth]{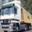}}\\
    \hspace{0.5pt}
    \subfigure[benign]{
        \includegraphics[width=0.17\linewidth]{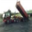}}
    \subfigure[adv]{
        \includegraphics[width=0.17\linewidth]{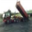}}
    \subfigure[CBDR]{
        \includegraphics[width=0.17\linewidth]{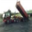}}
    \subfigure[SR]{
        \includegraphics[width=0.17\linewidth]{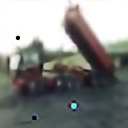}}
    \subfigure[NIP]{
        \includegraphics[width=0.17\linewidth]{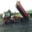}}
\caption{Visualizations of adversarial examples after defense on VGG19 model of CIFAR-10. Adversarial examples of these three lines are crafted by BIM, Boundary and PWA, respectively. ``adv" denotes adversarial examples. CBDR, SR and NIP are used as defense methods.}
\label{appendix-cifar-visual}
\end{figure}

\makeatletter

\begin{figure}[t]
\renewcommand{\@thesubfigure}{\hskip\subfiglabelskip}
\makeatother
\subfigbottomskip=-13pt
\setlength{\abovecaptionskip}{16pt}
\centering
    \subfigure[]{
        \includegraphics[width=0.17\linewidth]{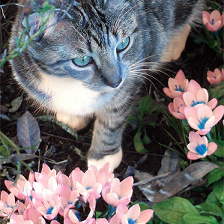}}
    \subfigure[]{
        \includegraphics[width=0.17\linewidth]{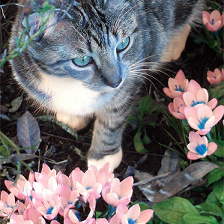}}
    \subfigure[]{
        \includegraphics[width=0.17\linewidth]{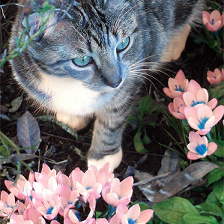}}
    \subfigure[]{
        \includegraphics[width=0.17\linewidth]{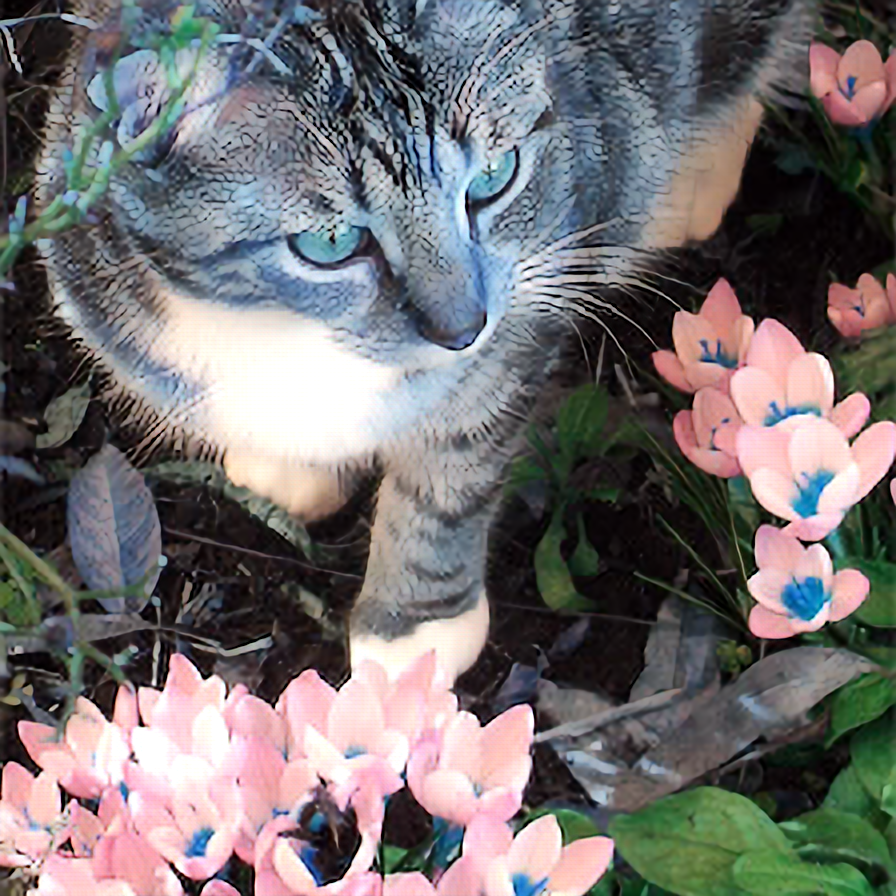}}
    \subfigure[]{
        \includegraphics[width=0.17\linewidth]{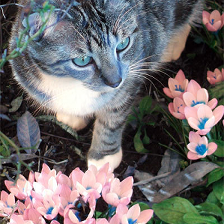}}\\
    \subfigure[]{
        \includegraphics[width=0.17\linewidth]{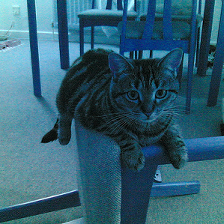}}
    \subfigure[]{
        \includegraphics[width=0.17\linewidth]{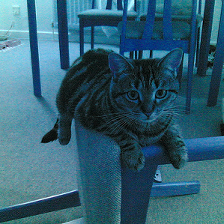}}
    \subfigure[]{
        \includegraphics[width=0.17\linewidth]{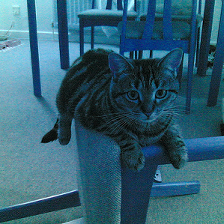}}
    \subfigure[]{
        \includegraphics[width=0.17\linewidth]{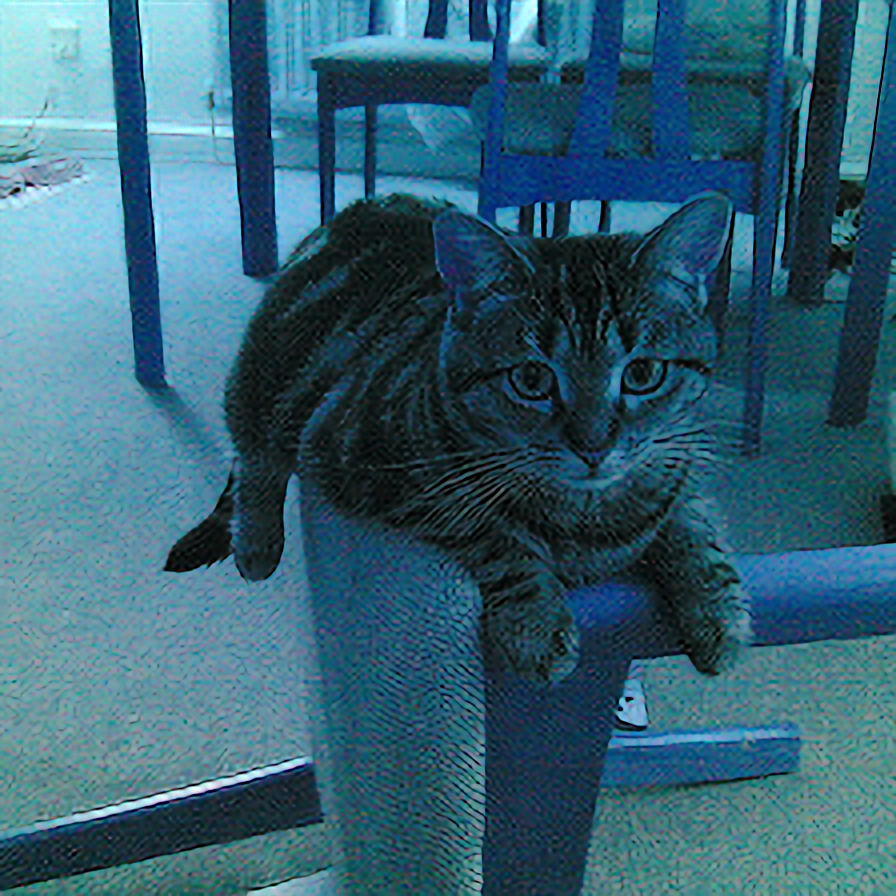}}
    \subfigure[]{
        \includegraphics[width=0.17\linewidth]{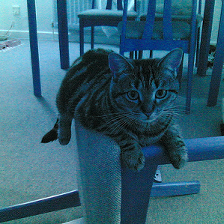}}\\
    \hspace{0.5pt}
    \subfigure[benign]{
        \includegraphics[width=0.17\linewidth]{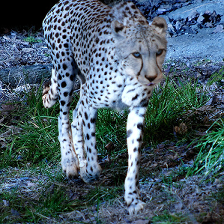}}
    \subfigure[adv]{
        \includegraphics[width=0.17\linewidth]{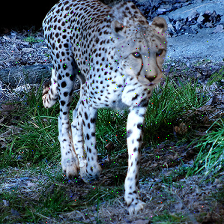}}
    \subfigure[CBDR]{
        \includegraphics[width=0.17\linewidth]{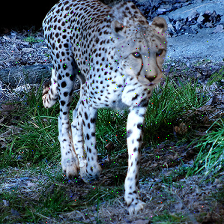}}
    \subfigure[SR]{
        \includegraphics[width=0.17\linewidth]{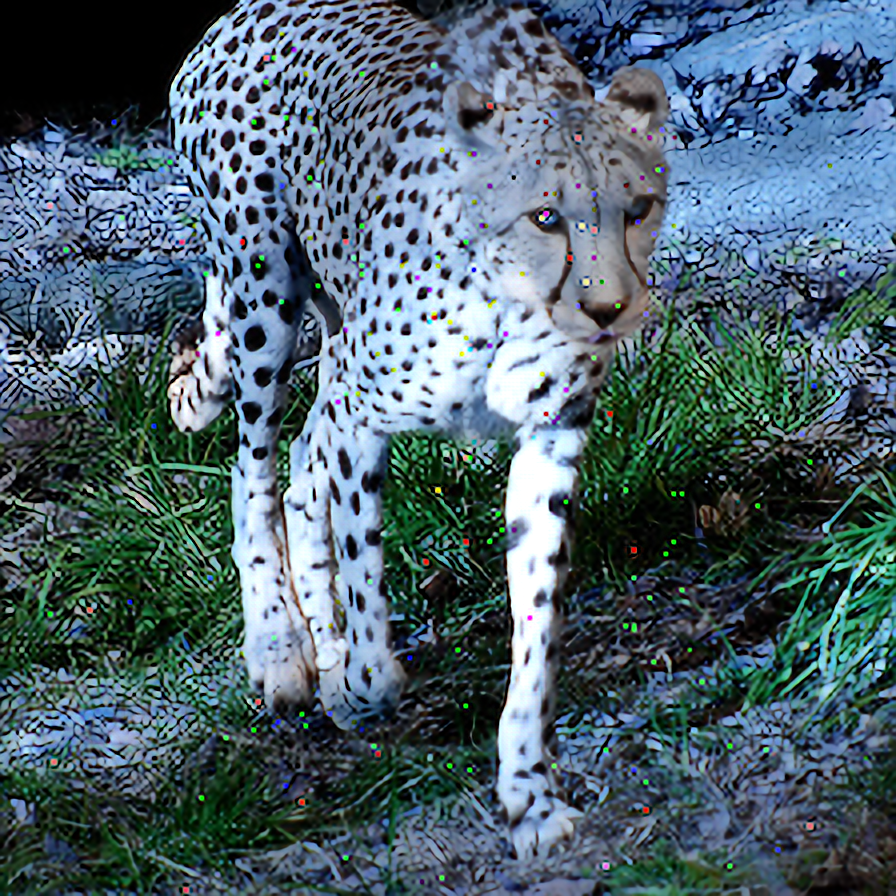}}
    \subfigure[NIP]{
        \includegraphics[width=0.17\linewidth]{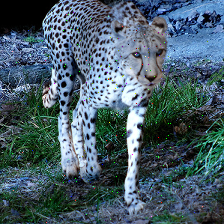}}
\caption{Visualizations of adversarial examples after defense on MobileNetV1 model of a-ImageNet. Adversarial examples of these three lines are crafted by BIM, Boundary and PWA, respectively. CBDR, SR and NIP are used as defense methods.}
\label{appendix-tiny-visual}
\vspace{-10pt}
\end{figure}

\subsection{Visualizations of Adversarial Examples After Defense\label{visual-after-def}}
Here we show visualizations of adversarial examples after defense on VGG19 model of CIFAR-10 and MobileNetV1 model of a-ImageNet, as shown in Fig. \ref{appendix-cifar-visual} and Fig. \ref{appendix-tiny-visual}, respectively. Adversarial examples of these three lines are crafted by BIM, Boundary and PWA. CBDR, SR and NIP are used as defense methods.

\subsection{Image Quality Analysis Based on PSNR}
We focus more on image quality after defense for interpretation.

\begin{figure}[htbp]
\centering
    \subfigure[CIFAR-10, VGG19]{
        \includegraphics[width=0.7\linewidth]{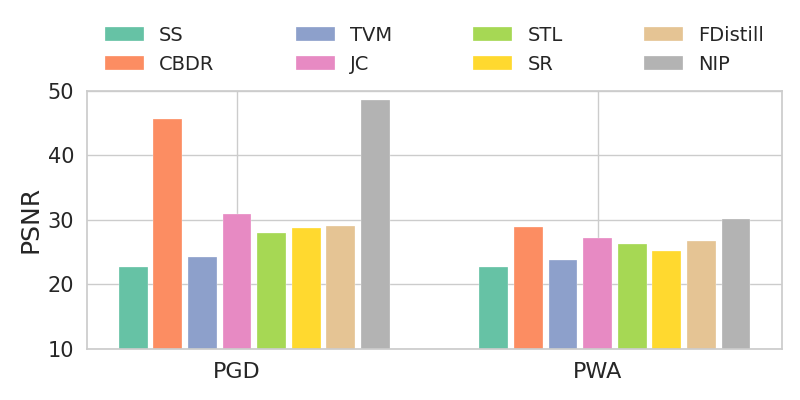}}\\
        \vspace{-5pt}
    \subfigure[a-ImageNet, MobileNetV1]{
        \includegraphics[width=0.7\linewidth]{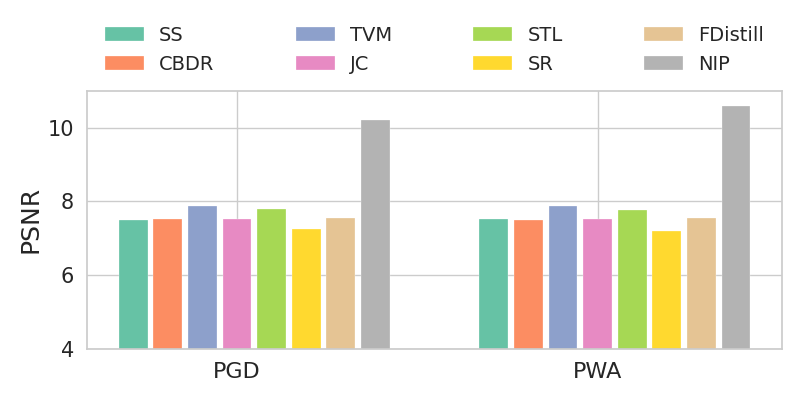}}
        \vspace{-5pt}
\caption{PSNR values of seven reactive defense baselines and NIP. }
\label{psnr}
\vspace{-10pt}
\end{figure}

\textbf{Implementation Details.} (1) 1,000 adversarial examples on VGG19 of CIFAR-10 and MobileNetV1 of a-ImageNet are used, generated by PGD and PWA. (2) For measurement, peak signal to noise ratio (PSNR) is adopted for further comparison with reactive baselines, where $\text{PSNR} = 20 \times \log_{10}{\frac{max_I^2}{\text{MSE}}} $. $\text{MSE}$ can be calculated as: $\text{MSE} = \frac{1}{m\times n} \sum_{i=0}^{m-1} \sum_{j=0}^{n-1} {[I-K]}^2$, where $I$ and $K$ are two images with the same size $m \times n$. $max_I^2$ is the possible maximum pixel value of image $I$. Images with larger PSNR have better image quality and smaller distortions. (3) Results are shown in Fig. \ref{psnr}. 

\textbf{Results and Analysis.} As shown in the Fig. \ref{psnr}, defense images generated by NIP provide better quality than baselines, with larger PSNR values ($ \sim \times 1.4$ on average) in all cases. It is obvious gray bars are the highest while bars of other colors are lower. By reflecting neuron influence to pixels, class-relevant pixels are strengthened while those may be maliciously manipulated are targeted and weakened. So NIP removes adversarial perturbations with useful details unharmed. On the contrary, baselines remove some important pixels together with adversarial perturbations during the transformation procedure. As a result, larger distortions and lower image quality are observed. Besides, PSNR of PWA is lower than that of PGD. Adversarial examples of PWA after defense show lower quality because perturbations are larger than PGD. 

\section{Details And Results of $\text{NIP}_n$}
We have introduced $\text{NIP}_n$ for comparison, to explain why we modify input values. Here we detail the defense process of $\text{NIP}_n$ and provide experiment results.

\begin{figure}[htbp]
\centering
        \includegraphics[width=0.92\linewidth]{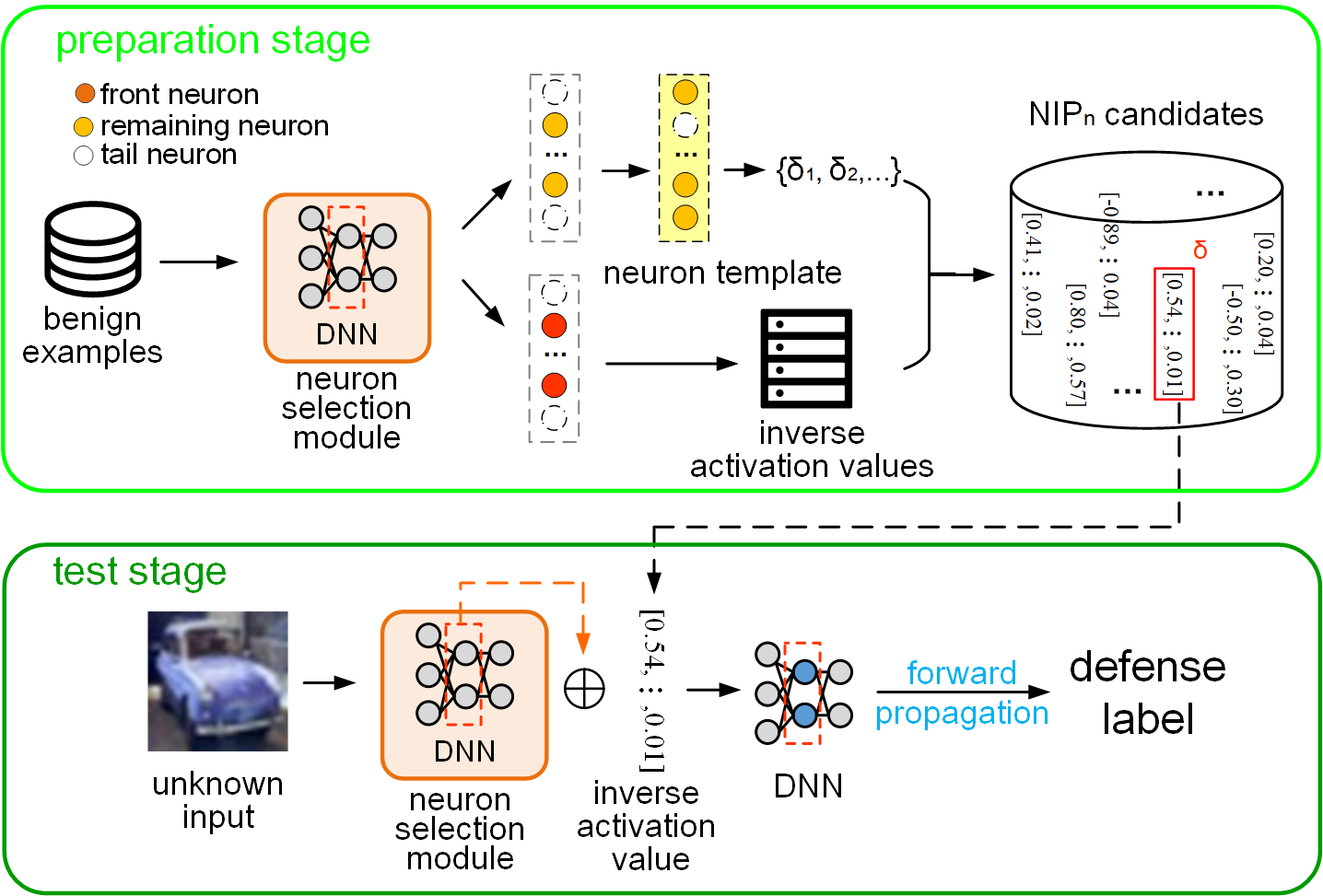} 
        \vspace{-5pt}
\caption{The overflow of $\text{NIP}_n$.}
	\label{nipn}
	\vspace{-5pt}
\end{figure}

The workflow of $\text{NIP}_n$ is shown in Fig. \ref{nipn}. Specifically, in the preparation stage, front, tail and remaining neurons are identified from benign examples. After that, front and tail neurons are used to generate inverse activation value. They are attached with their $\delta$ to construct $\text{NIP}_n$ candidates. In the test stage, the corresponding inverse activation value is found from $\text{NIP}_n$ candidates, and added to the chosen layer. When the model is fed with the unknown input, the activation value of the chosen layer will be modified, and then the defense label will be output after forward propagation.

We have conducted experiments for comparing NIP with $\text{NIP}_n$, including (1) defense effectiveness and (2) the parameter sensitivity on $\eta$. 

Following the experiment setting in the main experiment, we choose one model for each dataset: VGG19 of CIFAR-10, SqueezeNet~\cite{iandola2016squeezenet} of CIFAR-100~\cite{krizhevsky2009learning}, LeNet-5~\cite{lecun2015lenet} of GTSRB and MobileNetV1 of a-ImageNet. Global average pooling is chosen for SqueezeNet and MobileNetV1, while flatten layer is chosen for other models. For fair comparison of NIP in controlling front and tail neurons, $k_1$ and $k_2$ are both set to 5.

\subsection{Comparisons of Defense Effectiveness}
\begin{figure}[htbp]
\vspace{-5pt}
\centering
    \subfigure[Activation values]{
        \includegraphics[width=0.7\linewidth]{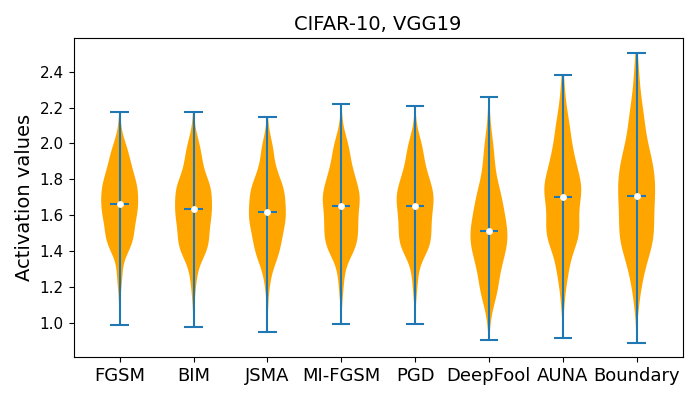}}\\
        \vspace{-5pt}
    \subfigure[Pixel values]{
        \includegraphics[width=0.7\linewidth]{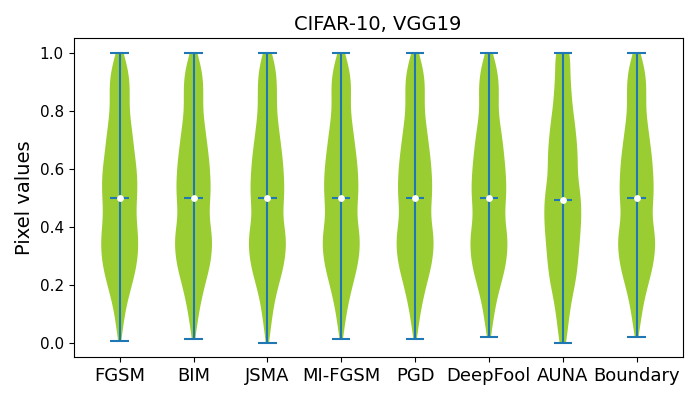}}
        \vspace{-5pt}
\caption{The distribution of neuron activation values and pixel values among different adversarial examples.}
\label{diff_nipn}
\vspace{-10pt}
\end{figure}

We calculate DSR on 1000 adversarial examples per attack. We set $\eta$=1 where NIP and $\text{NIP}_n$ show the best defense performance theoretically. Results against general attacks are shown in Table \ref{dsr_nipn}. We visualize activation values and pixel values on different adversarial examples. Average values of 100 adversarial examples per attack are used for calculation and results are shown in Fig. \ref{diff_nipn}. The maximum, minimum and median values are presented.

\begin{table}[htbp]
\centering
\caption{DSR between $\text{NIP}_n$ and NIP against general attacks.}
\resizebox{1\linewidth}{!}{
\begin{tabular}{cccccccccc}
\toprule
\multicolumn{2}{l}{\textbf{Datasets}} & \multicolumn{2}{c}{CIFAR-10} & \multicolumn{2}{c}{CIFAR-100}  & \multicolumn{2}{c}{GTSRB}   & \multicolumn{2}{c}{a-ImageNet}  \\ \cline{3-10} 
\multicolumn{2}{l}{\textbf{Models}}   & \multicolumn{2}{c}{VGG19}    & \multicolumn{2}{c}{SqueezeNet} & \multicolumn{2}{c}{LeNet-5} & \multicolumn{2}{c}{MobileNetV1} \\ \cline{3-10} 
\multicolumn{2}{l}{\textbf{Methods}} & NIP     & $\text{NIP}_n$    & NIP     & $\text{NIP}_n$    & NIP     & $\text{NIP}_n$    & NIP     & $\text{NIP}_n$    \\ \hline
\multirow{10}{*}{\textbf{\rotatebox{90}{Attacks}}} & FGSM   & \textbf{96.10\%}  & 94.80\%  & \textbf{99.80\%}    & 96.80\%  & \textbf{95.20\%}  & 91.20\% & \textbf{99.90\%}    & 96.90\%   \\
              & BIM                  & \textbf{99.20\%} & 97.70\% & 99.50\%          & 99.60\% & \textbf{99.50\%} & 91.40\% & \textbf{99.90\%} & 98.40\% \\
              & MI-FGSM              & \textbf{98.80\%} & 95.80\% & \textbf{99.80\%} & 99.40\% & \textbf{99.80\%} & 90.80\% & \textbf{99.90\%} & 97.60\% \\
              & JSMA                 & \textbf{98.70\%} & 97.60\% & 99.50\%          & 99.70\% & \textbf{98.40\%} & 85.80\% & \textbf{98.70\%} & 97.20\% \\
              & PGD                  & \textbf{99.30\%} & 96.10\% & \textbf{99.60\%} & 99.20\% & \textbf{99.60\%} & 91.70\% & \textbf{99.90\%} & 99.80\% \\
              & DeepFool             & \textbf{94.90\%} & 54.90\% & \textbf{99.30\%} & 98.20\% & \textbf{77.50\%} & 51.80\% & \textbf{99.90\%} & 99.60\% \\
              & UAP                  & \textbf{98.00\%} & 75.00\% & \textbf{96.60\%} & 40.10\% & \textbf{51.70\%} & 32.10\% & \textbf{99.30\%} & 43.50\% \\
              & AUNA                 & \textbf{95.50\%} & 34.50\% & \textbf{98.50\%} & 94.80\% & \textbf{78.80\%} & 67.40\% & \textbf{90.40\%} & 89.00\% \\
              & PWA                  & 99.30\%          & 99.50\% & \textbf{98.60\%} & 98.20\% & \textbf{98.90\%} & 93.80\% & \textbf{99.70\%} & 96.80\% \\
              & Boundary             & \textbf{99.50\%} & 10.20\% & \textbf{99.30\%} & 98.40\% & \textbf{99.90\%} & 95.00\% & \textbf{99.70\%} & 95.50\% \\ \bottomrule
\end{tabular}
}
\label{dsr_nipn}
\vspace{-10pt}
\end{table}

\begin{figure}[htbp]
\centering
        \includegraphics[width=0.5\linewidth]{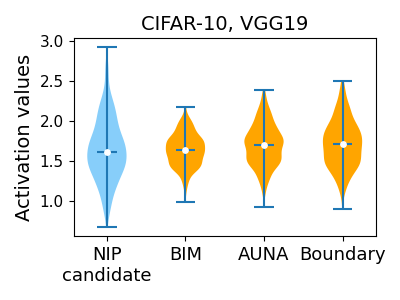} 
        \vspace{-5pt}
\caption{The distribution of NIP candidates on activation values.}
	\label{nip_ac}
	\vspace{-5pt}
\end{figure}

As shown in Table \ref{dsr_nipn}, $\text{NIP}_n$, which directly modifying neurons, shows inferiority than NIP in defense performance. We speculate the reason that activation values of different neurons are quite different among various attacks, so inverse activation values in $\text{NIP}_n$ candidates cannot maintain high defense effect on general attacks. Compared with neuron activation values, the values of adversarial perturbation show smaller difference among attacks, as shown in Fig. \ref{diff_nipn}. So inverse perturbations work better on various attacks than inverse activation values. 

We also visualize the neuron activation values of NIP candidates when they are added on adversarial examples in Fig. \ref{nip_ac}. As figure suggests, NIP candidates have larger bound in activation values than adversarial examples. So DSR of NIP is over 95\% on these adversarial attacks. Compared with $\text{NIP}_n$, NIP can cover more adversarial examples in activation values. So NIP is more effective in defense than $\text{NIP}_n$.

\subsection{Parameter Sensitivity on $\eta$}
We consider the situation that only a small number of NIP candidates are available for defense. We set $\eta$=0.3, 0.5 and 1 to study the impact of NIP candidates, and the results are shown in Table \ref{nipn_eta}. DSR is measured on 1000 PGD adversarial examples.

\begin{table}[htbp]
\centering
\caption{Comparison of DSR with different $\eta$.}
\resizebox{1\linewidth}{!}{
\begin{tabular}{cccccccc}
\toprule
\multirow{2}{*}{\textbf{Datasets}} & \multirow{2}{*}{\textbf{Models}} & \multicolumn{2}{c}{$\eta$=0.3} & \multicolumn{2}{c}{$\eta$=0.5} & \multicolumn{2}{c}{$\eta$=1} \\ \cline{3-8} 
           &             & NIP              & $\text{NIP}_n$    & NIP              & $\text{NIP}_n$    & NIP              & $\text{NIP}_n$    \\ \hline
CIFAR-10   & VGG19       & \textbf{90.70\%} & 72.80\% & \textbf{94.00\%} & 78.50\% & \textbf{99.30\%} & 96.10\% \\
a-ImageNet & MobileNetV1 & \textbf{93.50\%} & 75.60\% & \textbf{96.40\%} & 81.60\% & \textbf{99.90\%} & 99.80\% \\ \bottomrule
\end{tabular}
    }
    \label{nipn_eta}
    \vspace{-5pt}
\end{table}

NIP is still better than $\text{NIP}_n$ under different $\eta$. $\text{NIP}_n$ shows sharp decrease when $\eta$ is smaller than 1. This indicates that the defense performance of $\text{NIP}_n$ is quite sensitive to the number of NIP candidates. With regard to NIP, DSR of it is stable when fewer candidates are available. We speculate the reason that neuron activation values are quite different among different adversarial examples, which increases the difficulty of defense adaptivity. 
With the same number of candidates, modifying input is a more effective and general way for mitigating adversarial effect. So we choose to modify input values instead of modifying neurons.

In a word, NIP that modifies input values are better than the strategy of modifying neurons in the following two aspects: (1) better defense performance; (2) higher generality and low parameter sensitivity. So we choose to modify input values rather than modify neurons in the test stage.

\section{Neuron Template And Labels}
From the definition of neuron influence, it is correlated with labels. But we create only one neuron template for each model, which assumes the template is label agnostic. This may be confusing to some readers. Here we provide experiments and analysis on correlation between neuron template and labels.

Templates of different classes have something in common and neuron template is generated by taking intersection of remaining neurons from different classes. We calculate the similarity $\delta$ between remaining neurons of the given input and neuron template, which can be used to measure the correlation between the input and its corresponding label. That is, for one coming input, we can use similarity $\delta$ calculated by template, to find its corresponding candidate of the label.
Therefore, neuron template is related with labels and the similarity $\delta$ can reflect the relevance between template and labels.

\begin{figure}[htbp]
\vspace{-5pt}
\centering
        \includegraphics[width=0.8\linewidth]{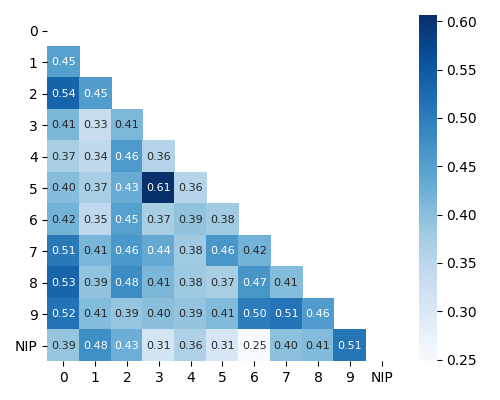} 
        \vspace{-5pt}
\caption{Visualizations of the similarity between neuron template from different classes.}
	\label{template_sim}
	\vspace{-10pt}
\end{figure}

We have conducted experiments on VGG19 of CIFAR-10 to verify our assumption. We use benign examples from one class to create neuron template and further compare the similarity of neuron templates from different classes. Intersection over union (IoU) of neurons in neuron template calculated from different classes is calculated to measure the consistency of them. In particular, a greater value denotes higher consistency. The visualization of the similarity between NIP’s neuron template and neuron templates from different class via heatmaps is shown in Fig. \ref{template_sim}. As the color goes deeper, the value increases. Numbers denote neuron template from different class. ``NIP” denotes neuron template used by NIP. Here, we set $k$=5.

As Fig. \ref{template_sim} suggests, neuron template calculated by benign examples from different classes show something in common but they are still different with each other. Meanwhile, these templates have similar neurons as NIP's neuron template. NIP uses a batch of randomly-chosen benign examples to calculate neuron template, which contains features from remaining neurons in different classes. Therefore, consistent with our assumption, neuron template is correlated with labels.

\section{Comparisons Between Neuron-level Defenses}
We have conducted experiments and made comparisons about neuron-level defense methods for evaluation, including SAP~\cite{DhillonALBKKA18}, NS~\cite{zhang2019neuron}, Suri et al.~\cite{suri2020one}, CAS~\cite{bai2020improving} and SNS~\cite{zhang2020interpreting}. Experiments are implemented in three aspects: (1) defense effectiveness; (2) defense efficiency; (3) comparisons of selected neurons. 

\subsection{Defense Effectiveness And Efficiency}
Experiments are conducted on VGG19 of CIFAR-10, SqueezeNet of CIFAR-100 and ResNet20 of GTSRB. We calculate defense success rate (DSR) on 1000 adversarial examples per attack, following the setting in the main paper. Besides, we also calculate the total time of handling 1000 adversarial examples for 3 times as the average time. Results are shown in Table \ref{neuron level DSR}. NS fails to handle multi-class dataset as CIFAR-100 so the experiment is not conducted.

\begin{table}[htbp]
\centering
\caption{DSR and running time comparison between NIP and neuron-level defenses.}
\resizebox{1\linewidth}{!}{
\Large
\begin{tabular}{ccccccc}
\toprule
\multirow{2}{*}{\textbf{Datasets}} &
  \multirow{2}{*}{\textbf{Models}} &
  \multirow{2}{*}{\textbf{Methods}} &
  \multicolumn{3}{c}{\textbf{DSR}} &
  \multirow{2}{*}{\textbf{Time/s}} \\ \cline{4-6}
                           &                             &             & FGSM             & PGD               & PWA              &                 \\ \hline
\multirow{6}{*}{CIFAR-10}  & \multirow{6}{*}{VGG19}      & SAP         & 74.50\%          & 78.60\%           & 62.60\%          & 4350.12         \\
                           &                             & NS          & 30.30\%          & 35.20\%           & 23.10\%          & 50347.20        \\
                           &                             & Suri et al. & 67.30\%          & 68.10\%           & 55.20\%          & 1260.39         \\
                           &                             & CAS         & 89.90\%          & 96.20\%           & 95.70\%          & 1005.61         \\
                           &                             & SNS         & 80.20\%          & 90.10\%           & 80.10\%          & 796.40          \\
                           &                             & NIP         & \textbf{96.10\%} & \textbf{99.30\%}  & \textbf{99.30\%} & \textbf{39.50}  \\ \hline
\multirow{6}{*}{CIFAR-100} & \multirow{6}{*}{SqueezeNet} & SAP         & 68.20\%          & 69.30\%           & 54.60\%          & 4420.31         \\
                           &                             & NS          & /                & /                 & /                & /               \\
                           &                             & Suri et al. & 59.30\%          & 62.00\%           & 53.70\%          & 1596.84         \\
                           &                             & CAS         & 72.50\%          & 73.40\%           & 61.50\%          & 1395.02         \\
                           &                             & SNS         & 85.40\%          & 86.70\%           & 82.50\%          & 821.63          \\
                           &                             & NIP         & \textbf{99.80\%} & \textbf{100.00\%} & \textbf{98.60\%} & \textbf{214.30} \\ \hline
\multirow{6}{*}{GTSRB}     & \multirow{6}{*}{ResNet20}   & SAP         & 72.50\%          & 73.10\%           & 68.30\%          & 3690.69         \\
                           &                             & NS          & 30.80\%          & 32.40\%           & 25.10\%          & 136983.67       \\
                           &                             & Suri et al. & 68.90\%          & 69.00\%           & 53.60\%          & 1395.35         \\
                           &                             & CAS         & 87.30\%          & 94.20\%           & 69.60\%          & 1298.20         \\
                           &                             & SNS         & 88.40\%          & 90.70\%           & 60.80\%          & 805.63          \\
                           &                             & NIP         & \textbf{99.20\%} & \textbf{99.50\%}  & \textbf{98.70\%} & \textbf{60.71}  \\ \bottomrule
\end{tabular}}
    \label{neuron level DSR}
\end{table}

As observed, NIP outperforms other neuron-level defense methods with a considerable margin. It obtains highest DSR against various attacks, with the shortest running time. By enhancing the most class-relevant neurons and suppressing irrelevant neurons that may be utilized by general attacks in one layer, the effect of adversarial perturbations can be well alleviated. Instead of directly modifying neurons in the targeted model, we choose to modify the input. It saves much time on retraining or constructing neuron selectors. So NIP is empirically a more light-weighted and efficient solution against various attacks.

\subsection{Comparisons of Selected Neurons}
\begin{figure}[htbp]
\centering
        \includegraphics[width=0.92\linewidth]{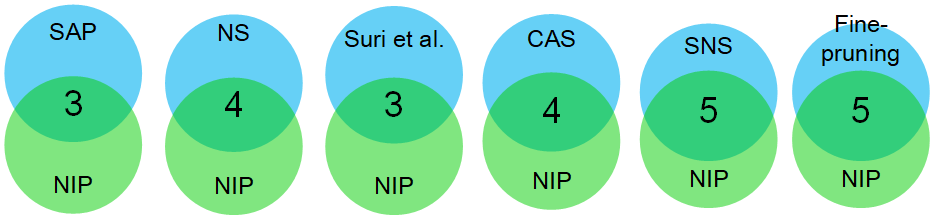} 
        \vspace{-5pt}
\caption{The overlap of selected neurons used in the defense.}
	\label{appendix iou}
	\vspace{-10pt}
\end{figure}

We further compare the overlap of 10 selected neurons activated by NIP and these neuron-level defense on VGG19 of CIFAR-10. 100 adversarial examples crafted by PGD are used. For each defense, 10 neurons are selected in total. Specifically, top-10 neurons are chosen for baselines, while top-5 and bottom-5 are selected for NIP. We calculate top-10 neurons in the flatten layer or the activation layer. Results are shown in Fig. \ref{appendix iou}, where each circle denotes 10 neurons. Numbers attached denote the number of overlapped selected neurons on average.

Neurons selected by NIP show difference among neuron-level defenses. Based on randomness, neurons that SAP chooses show large difference with NIP, which can be attributed to stochastic strategy. As for NS, on large models with complex structures, the neuron selection strategy may be inaccurate and invalid. Meanwhile, neurons selected by it show less overlap with NIP. As for retraining strategies like CAS and SNS, they identify neurons according to the predicted label or neuron activation values due to perturbations, so more common neurons are selected with NIP. 

To summarize, NIP shows superior defense capability than other neuron-level defenses, in the following aspects: (1) better DSR on adversarial examples; (2) less time complexity. Neurons selected by NIP show difference with that of baselines, which indicates the novelty of our selection strategy.

\bibliographystyle{IEEEtran}
\bibliography{ref}